\documentclass[10pt,twocolumn,letterpaper]{article}

\pdfoutput=1
\usepackage{cvpr}
\usepackage{times}
\usepackage{epsfig}
\usepackage{graphicx}
\usepackage{amsmath}
\usepackage{amssymb}
\usepackage{subfigure}
\usepackage[linesnumbered,lined,vlined,ruled,commentsnumbered]{algorithm2e}
\usepackage{mathtools}
\usepackage{verbatim}
\usepackage{booktabs} 
\usepackage{multirow}
\usepackage{array}
\usepackage{footnote}
\usepackage{tabularx}
\usepackage{cite}

\usepackage[pagebackref=true,breaklinks=true,letterpaper=true,colorlinks,bookmarks=false]{hyperref}
\usepackage{amsmath,amsfonts,bm}

\def\eqref#1{equation~\ref{#1}}
\def\1{\bm{1}}

\DeclareMathAlphabet{\mathsfit}{\encodingdefault}{\sfdefault}{m}{sl}
\SetMathAlphabet{\mathsfit}{bold}{\encodingdefault}{\sfdefault}{bx}{n}

\DeclareMathOperator*{\argmin}{arg\,min}

\newcommand{\myparagraph}[1]{\vspace{0.1em}\noindent\textbf{#1}}
\newcommand{\myparagraphsupp}[1]{\vspace{0.1em}\noindent{\textcolor{red}{#1}}}
\newcommand{\mycaptionsupp}[1]{{\textcolor{red}{#1}}}
\newcommand{\redt}[1]{\textcolor[rgb]{0,0,0}{#1}}
\newcommand{\updatedredt}[1]{\textcolor[rgb]{0,0,0}{#1}}

\newcommand{\cotronlvsapce}{\vspace{-0.0cm}}
\newcommand{\cotronlcaptionvsapce}{\vspace{0.2cm}}
\cvprfinalcopy
\def\cvprPaperID{} 

\ifcvprfinal\fi
\begin{document}

\title{
Mnemonics Training: Multi-Class Incremental Learning without Forgetting}
\author{Yaoyao Liu$^{1,2}\thanks{This work was done during Yaoyao's internship supervised by Qianru.}$ 
\quad Yuting Su$^{1}\thanks{Corresponding authors.}$ 
\quad An-An Liu$^{1}\footnotemark[2]$
\quad Bernt Schiele$^{2}$ 
\quad Qianru Sun$^{3}$\\
\\
\small  $^{1}$School of Electrical and Information Engineering, Tianjin University\\
\small  $^{2}$Max Planck Institute for Informatics, Saarland Informatics Campus\\
\small  $^{3}$School of Computing and Information Systems, Singapore Management University\\
%\small  {\texttt{\{yaoyao.liu, schiele, qsun\}@mpi-inf.mpg.de}} \\
%\small  {\texttt{\{liuyaoyao, ytsu, liuanan\}@tju.edu.cn}} \quad  {\texttt{qianrusun@smu.edu.sg}}
}

\maketitle
\thispagestyle{empty}

\newcommand{\beginsupp}{%
        \setcounter{table}{0}
        \renewcommand{\thetable}{S\arabic{table}}%
        \setcounter{figure}{0}
        \renewcommand{\thefigure}{S\arabic{figure}}%
     }

%%%%%%%%% ABSTRACT
\begin{abstract}
Multi-Class Incremental Learning (MCIL) aims to learn new concepts by incrementally updating a model trained on previous concepts. However,
there is an inherent trade-off to effectively learning new concepts 
without catastrophic forgetting of previous ones.
To alleviate this issue, it has been proposed to keep around a few examples of the previous concepts but the effectiveness of this approach heavily depends on the representativeness of these examples. 
This paper proposes a novel and automatic framework we call \emph{mnemonics}, where we parameterize exemplars and make them optimizable in an end-to-end manner.
We train the framework through bilevel optimizations, i.e., model-level and exemplar-level.
We conduct extensive experiments on three MCIL benchmarks, CIFAR-100, ImageNet-Subset and ImageNet, and show that using \emph{mnemonics} exemplars 
can surpass the state-of-the-art by a large margin. 
Interestingly and quite intriguingly, the \emph{mnemonics} exemplars tend to be on the boundaries between different classes\footnote{Code: \href{https://github.com/yaoyao-liu/mnemonics-training}{https://github.com/yaoyao-liu/mnemonics-training}}.
\end{abstract}

%%%%%%%%% BODY TEXT
\section{Introduction}
\label{sec_introduction}

Natural learning systems such as humans inherently work in an incremental manner as the number of concepts increases over time. 
They naturally learn new concepts while not forgetting previous ones.
In contrast, current machine learning systems, when continuously updated using novel incoming data, suffer from catastrophic forgetting (or catastrophic interference), as the updates can override knowledge acquired from previous data~\cite{mccloskey1989catastrophic, McRae1993Catastrophic, Ratcliff1990catastrophic, shin2017continual, KemkerMAHK18}.
This is especially true for multi-class incremental learning (MCIL) where one cannot replay all previous inputs. Catastrophic forgetting, therefore, becomes a major problem for MCIL systems.

\begin{figure}[t]
\centering
\includegraphics[width=3.1in]{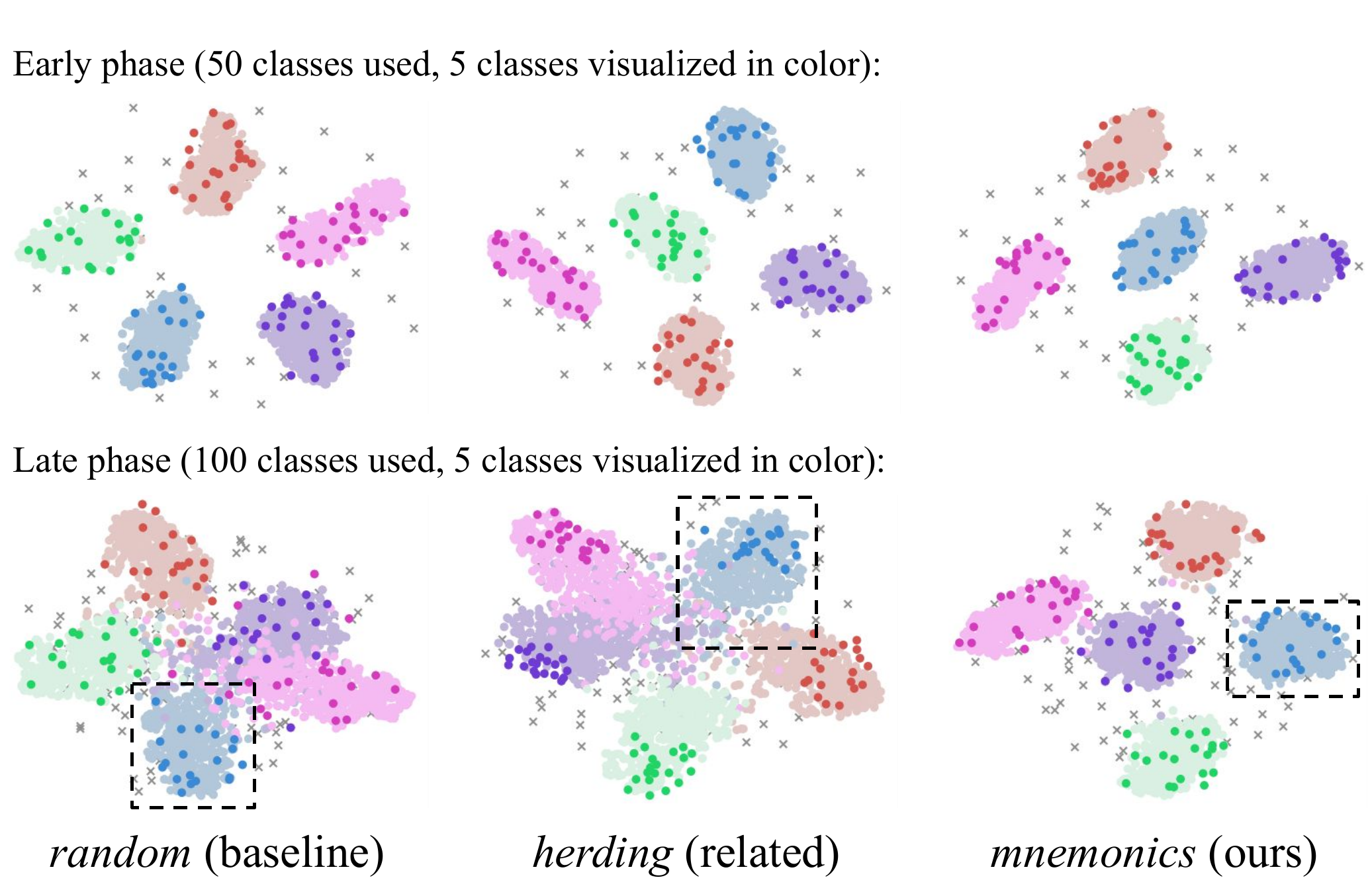}
\cotronlcaptionvsapce
\caption{The t-SNE~\cite{maaten2008visualizing}  results of three exemplar methods in two phases. The original data of $5$ colored classes occur in the early phase.
In each colored class, deep-color points are exemplars, and light-color ones show the original data as reference of the real data distribution. Gray crosses represent other participating classes, and each cross for one class. We have two main observations. (1)~Our approach results in much clearer separation in the data, than \emph{random} (where exemplars are randomly sampled in the early phase) and \emph{herding} (where exemplars are nearest neighbors of the mean sample in the early phase)~\cite{rebuffi2017icarl, hou2019learning, Wu2019LargeScale, Castro18EndToEnd}. (2)~Our learned exemplars mostly locate on the boundaries between classes.}
\label{fig_teaser}
\cotronlvsapce
%\vspace{-0.2cm}
\end{figure}

Motivated by this, a number of works have recently emerged~\cite{rebuffi2017icarl, Li18LWF, hou2019learning, Wu2019LargeScale, Castro18EndToEnd,li2019online}. 
Rebuffi et al.~\cite{rebuffi2017icarl} firstly defined a protocol for evaluating MCIL methods, i.e., to tackle the image classification task where the training data for different classes comes in sequential training phases.
As it is neither desirable nor scaleable to retain all data from previous concepts, in their protocol, they restrict the number of exemplars that can be kept around per class, e.g., only $20$ exemplars per class can be stored and passed to the subsequent training phases.
These ``$20$ exemplars'' are important to MCIL as they are the key resource for the model to refresh its previous knowledge. 
Existing methods to extract exemplars are based on heuristically designed rules, e.g., nearest neighbors around the average sample in each class (named \emph{herding}~\cite{welling2009herding})~\cite{rebuffi2017icarl, hou2019learning, Wu2019LargeScale, Castro18EndToEnd}, but turn out to be not particularly effective.
For example, iCaRL~\cite{rebuffi2017icarl} with \emph{herding} sees an accuracy drop of around $25\%$ in predicting $50$ previous classes in the last phase (when the number of classes increases to $100$) on CIFAR-100, compared to the upper-bound performance of using all examples. 
A t-SNE visualization of \emph{herding} exemplars is given in Figure~\ref{fig_teaser}, and shows that the separation between classes becomes weaker in later training phases.

In this work, we address this issue by developing an automatic exemplar extraction framework called \emph{mnemonics} where we parameterize the exemplars using image-size parameters, and then optimize them in an end-to-end scheme. 
Using \emph{mnemonics}, the MCIL model in each phase can not only learn the optimal exemplars from the new class data, but also adjust the exemplars of previous phases to fit the current data distribution. 
As demonstrated in Figure~\ref{fig_teaser}, \emph{mnemonics} exemplars yield consistently clear separations among classes, from early to late phases. 
When inspecting individual classes (as e.g. denoted by the black dotted frames in Figure~\ref{fig_teaser} for the ``blue'' class), we observe that the \emph{mnemonics} exemplars (dark blue dots) are mostly located on the boundary of the class data distribution (light blue dots), which is essential to derive high-quality classifiers.

Technically, \emph{mnemonics} has two models to optimize, i.e., the conventional model and the parameterized \emph{mnemonics} exemplars.
The two are not independent and can not be jointly optimized, as the exemplars learned in the current phase will act as the input data of later-phase models.
We address this issue using a bilevel optimization program (BOP)~\cite{SinhaMD18bilevelreview, mackay2019self}
that alternates the learning of two levels of models. We iterate this optimization through the entire incremental training phases.
In particular, for each single phase, we perform a local BOP that aims to distill the knowledge of new class data into the exemplars.
First, a temporary model is trained with exemplars as input. Then,
a validation loss on new class data is computed and the gradients are back-propagated to optimize the input layer, i.e., the parameters of the \emph{mnemonics} exemplars.
Iterating these two steps allows to derive representative exemplars for later training phases.
To evaluate the proposed \emph{mnemonics} method, we conduct extensive experiments for FOUR different baseline architectures
and 
on THREE MCIL benchmarks -- CIFAR-100, ImageNet-Subset and ImageNet. 
Our results reveal that \emph{mnemonics} consistently achieves top performance compared to baselines, e.g., $17.9\%$ and $4.4\%$ higher than \emph{herding}-based iCaRL~\cite{rebuffi2017icarl} and LUCIR~\cite{hou2019learning}, respectively, in the $25$-phase setting on the ImageNet~\cite{rebuffi2017icarl}.

\textbf{Our contributions} include:
(1)~A novel \emph{mnemonics} training framework that alternates the learning of exemplars and models in a global bilevel optimization program, where bilevel includes \emph{model-level} and \emph{exemplar-level};
(2)~A novel local bilevel optimization program (including \emph{meta-level} and \emph{base-level}) that trains exemplars for new classes as well as adjusts exemplars of old classes in an end-to-end manner;
(3)~In-depth experiments, visualization and explanation of \emph{mnemonics} exemplars in the feature space.
\section{Related Work}
\label{sec_related_works}

\myparagraph{Incremental learning} has a long history in machine learning~\cite{cauwenberghs2001incremental, Mensink13DistanceBased, Kuzborskij13MulticlassTransfer}. 
A uniform setting is that the data of different classes gradually come.
Recent works are either in the multi-task setting (classes from different datasets)~\cite{Li18LWF,shin2017continual,hu2018overcoming,chaudhry2018efficient,riemer2018learning}, or in the multi-class setting (classes from the identical dataset)~\cite{rebuffi2017icarl,hou2019learning,Wu2019LargeScale,Castro18EndToEnd}. 
Our work is conducted on the benchmarks of the latter one called multi-class incremental learning (MCIL).

A classic baseline method is called knowledge distillation using a transfer set~\cite{HintonVD15}, first applied to incremental learning by Li et al.~\cite{Li18LWF}. 
Rebuffi et al.~\cite{rebuffi2017icarl} combined this idea with representation learning, for which a handful of \emph{herding} exemplars are stored for replaying old knowledge. \emph{Herding}~\cite{welling2009herding} picks the nearest neighbors of the average sample per class~\cite{rebuffi2017icarl}.
With the same \emph{herding} exemplars,
Castro et al.~\cite{Castro18EndToEnd} tried a balanced fine-tuning and temporary distillation to build an end-to-end framework;
Wu et al.~\cite{Wu2019LargeScale} proposed a bias correction approach; and
Hou et al.~\cite{hou2019learning} introduced multiple techniques also to balance classifiers. 
Our approach is closely related to these works.
The difference lies in the way of generating exemplars. In the proposed \emph{mnemonics} training framework, the exemplars are optimizable and updatable in an end-to-end manner, thus more effective than previous ones.

Using synthesizing exemplars
is another solution that ``stores'' the old knowledge in generative models.
Related methods~\cite{shin2017continual, kamra2017deep, venkatesan2017strategy} used Generative Adversarial Networks (GAN)~\cite{goodfellow2014generative} to generate old samples in each new phase for data replaying, and good results were obtained in the multi-task incremental setting. However, their performance strongly depends on the GAN models which are notoriously hard to train. Moreover, storing GAN models requires memory, so these methods might not be  applicable to MCIL with a strict memory budget. 
Our \emph{mnemonics} exemplars are optimizable, and can be regarded as synthesized, while our approach is based on the direct parameterization of exemplars without training extra models.

\begin{figure*}
\centering
\includegraphics[width=6.8in]{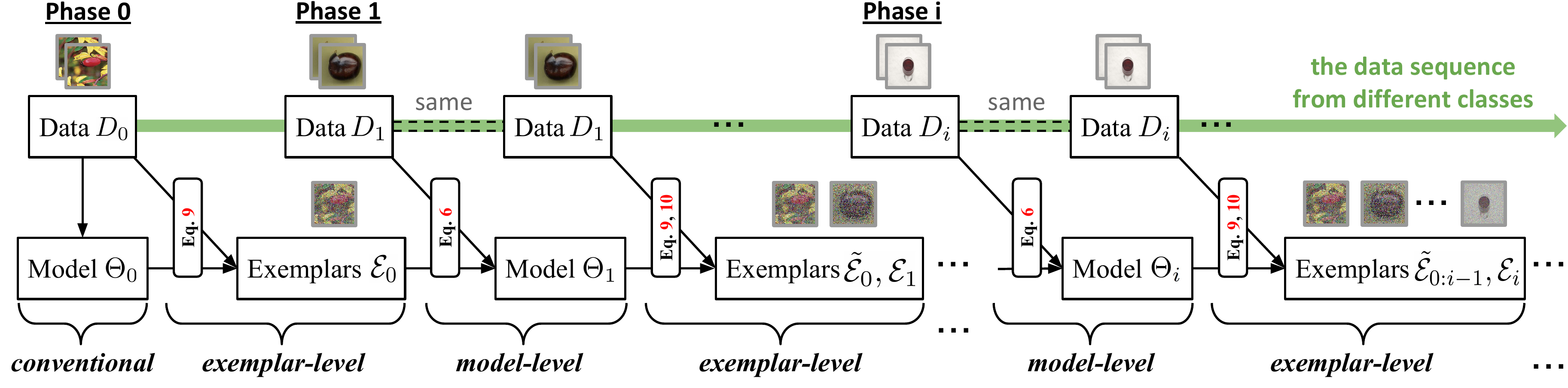}
\cotronlcaptionvsapce
\caption{
The computing flow of the proposed \emph{mnemonics} training. It is a global BOP that alternates the learning of \emph{mnemonics} exemplars (we call \emph{exemplar-level} optimization) and MCIL models (\emph{model-level} optimization). The \emph{exemplar-level} optimization within each phase is detailed in Figure~\ref{fig_framework_details}. $\tilde{\mathcal{E}}$ denotes the old exemplars adjusted to the current phase.}
\label{fig_framework}
\cotronlvsapce
\end{figure*}

\myparagraph{Bilevel optimization program (BOP)}~\cite{Wang2018Distillation,stackelberg1952theory,goodfellow2014generative} aims to solve two levels of problems in one framework where the A-level problem is the constraint to solve the B-level problem. 
It can be traced back to the Stackelberg competition~\cite{stackelberg1952theory} in the area of game theory.
Nowadays, 
it is widely applied in the area of machine learning. For instance, Training GANs~\cite{goodfellow2014generative} can be formulated as a BOP with two optimization problems: maximizing the reality score of generated images and minimizing the real-fake classification loss. 
Meta-learning~\cite{finn2017model,sun2019meta,zhang2019canet,li2019learning,zhang2020deepemd,sun2019mtlj} is another BOP in which a meta-learner is optimized subject to the optimality of the base-learner. 
Recently, MacKay et al.~\cite{mackay2019self} formulated the hyperparameter optimization as a BOP where the optimal model parameters in a certain time phase depend on hyperparameters, and vice versa. 
In this work, we introduce a global BOP that alternatively optimizes the parameters of the MCIL models and the \emph{mnemonics} exemplars across all phases. Inside each phase, we exploit a local BOP to learn (or adjust) the \emph{mnemonics} exemplars specific to the new class (or the previous classes).

\section{Preliminaries}
\label{sec_preli}

\myparagraph{Multi-Class Incremental Learning (MCIL)} was proposed in~\cite{rebuffi2017icarl} to evaluate classification models incrementally learned using a sequence of data from different classes. Its uniform setting is used in related works~\cite{rebuffi2017icarl,hou2019learning,Wu2019LargeScale,Castro18EndToEnd}.
It is different from the conventional classification setting, where training data for all classes are available from the start, in three aspects: (i) the training data come in as a stream where the sample of different classes occur in different time phases; (ii) in each phase, MCIL classifiers are expected to provide a competitive performance for all seen classes so far; and (iii) the machine memory is limited (or at least grows slowly), so it is impossible to save all data to replay the network training.

\myparagraph{Denotations.}
Assume there are $N+1$ phases (i.e, $1$ initial phase and $N$ incremental phases) in the MCIL system.
In the initial (the $0$-th) phase, we learn the model $\Theta_0$ on data $D_0$ using a conventional classification loss, e.g. cross-entropy loss,
and then save $\Theta_0$ to the memory of the system.
Due to the memory limitation, we can not keep the entire $D_0$, but instead we select and store a handful of exemplars $\mathcal{E}_0$ (evenly for all classes) as a replacement of $D_0$ with $|\mathcal{E}_0|\ll |D_0|$.
In the $i$-th incremental phase, we denote the previous exemplars $\mathcal{E}_0\sim \mathcal{E}_{i-1}$ shortly as $\mathcal{E}_{0:i-1}$.
We load $\Theta_{i-1}$ and $\mathcal{E}_{0:i-1}$ from the memory, and then use $\mathcal{E}_{0:i-1}$ and the new class data $D_i$ to train $\Theta_i$ initialized by $\Theta_{i-1}$.
During training, we use a classification loss and an MCIL-specific distillation loss~\cite{Li18LWF, rebuffi2017icarl}.
After each phase the model is evaluated on unseen data for all classes observed by the system so far.
We report the average accuracy over all $N+1$ phases as the final evaluation, following~\cite{rebuffi2017icarl,Wu2019LargeScale,hou2019learning}.

\myparagraph{Distillation Loss and Classification Loss.} Distillation Loss was originally proposed in~\cite{HintonVD15} and was applied to MCIL in~\cite{Li18LWF, rebuffi2017icarl}. 
It encourages the new $\Theta_i$ and previous $\Theta_{i-1}$ to maintain the same prediction ability on old classes.
Assume there are $K$ classes in $D_{0:i-1}$. Let $x$ be an image in $D_i$. $\hat{p}_{k}(x)$ and $p_k(x)$ denote the prediction logits of the $k$-th class from $\Theta_{i-1}$ and $\Theta_i$, respectively.
The distillation loss is formulated as
\begin{subequations}
\label{eq_distillation_loss}
\begin{equation}\label{eq_distillation_loss_1}
    \mathcal{L}_{d}(\Theta_i; \Theta_{i-1}; x) = -\sum_{k=1}^K\hat{\pi}_k(x)\text{log}\pi_k(x),
\end{equation}
\begin{equation}\label{eq_distillation_loss_2}
    \hat{\pi}_k(x)=\frac{e^{\hat{p}_k(x)/\tau}}{\sum_{j=1}^Ke^{\hat{p}_j(x)/\tau}}, \quad \pi_k(x)=\frac{e^{p_k(x)/\tau}}{\sum_{j=1}^Ke^{p_j(x)/\tau}},
\end{equation}
\end{subequations}
where $\tau$ is a temperature scalar set to be greater than $1$ to assign larger weights to smaller values.

We use the softmax cross entropy loss as the Classification Loss $\mathcal{L}_{c}$. Assume there are $M$ classes in $D_{0:i},$. This loss is formulated as
\begin{equation}\label{eq_classification_loss}
    \mathcal{L}_{c}(\Theta_i; x) = -\sum_{k=1}^{K+M}\delta_{y=k}\text{log}p_k(x),
\end{equation}
where $y$ is the ground truth label of $x$, and $\delta_{y=k}$ is an indicator function. 

\section{Mnemonics Training}
\label{sec_method}
\begin{figure*}[t]
\centering
\includegraphics[width=6.8in]{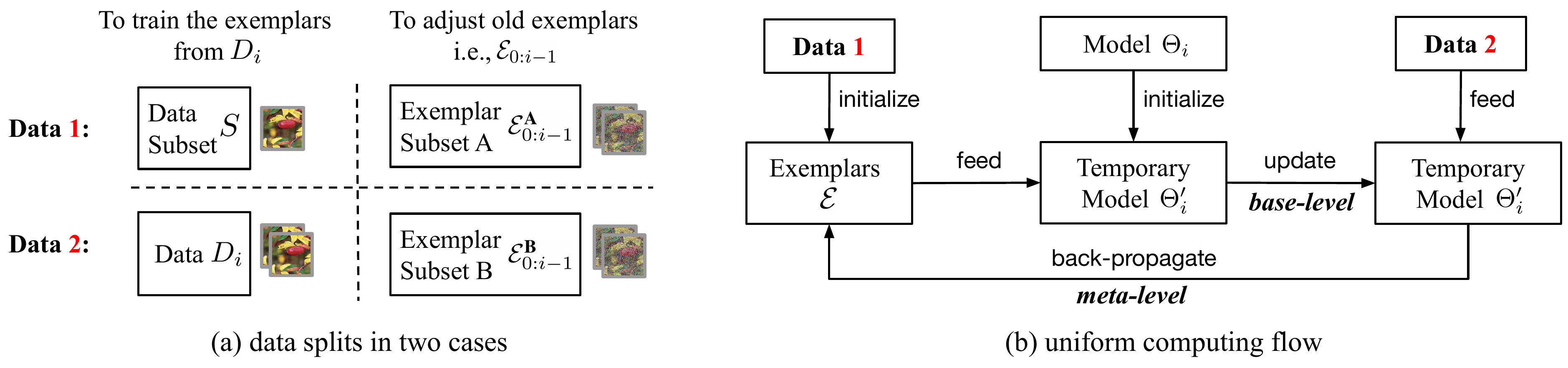}
\cotronlcaptionvsapce
\caption{The proposed local BOP framework that uses a uniform computing flow in (b) to handle two cases of \emph{exemplar-level} learning: training new class exemplars $\mathcal{E}_i$ from $D_i$; and adjusting old exemplars $\mathcal{E}_{0:i-1}$, with the data respectively given in (a). Note that (1) $\mathcal{E}^{\textbf{A}}_{0:i-1}$ and $\mathcal{E}^{\textbf{B}}_{0:i-1}$ are used as the validation set alternately for each other when adjusting $\mathcal{E}_{0:i-1}$; (2) $\mathcal{E}$ in (b) denote the  \emph{mnemonics} exemplars which are $\mathcal{E}_i$, $\mathcal{E}^{\textbf{A}}_{0:i-1}$, and $\mathcal{E}^{\textbf{B}}_{0:i-1}$
in Eq.~\ref{eq_update_ei_sgd},~\ref{eq_update_e_sgdAB} and ~\ref{eq_update_e_sgdBA}, respectively. 
}
\label{fig_framework_details}
\cotronlvsapce
\end{figure*}

As illustrated in Figure~\ref{fig_framework}, the proposed \emph{mnemonics} training alternates the learning of classification models and \emph{mnemonics} exemplars across all phases, where \emph{mnemonics} exemplars are not just data samples but can be optimized and adjusted online.
We formulate this alternative learning with a global \emph{Bilevel Optimization Program (BOP)} composed of \emph{model-level} and \emph{exemplar-level} problems (Section~\ref{subsec_global_training_flow}), and provide the solutions in Section~\ref{subsec_model_level} and Section~\ref{subsect_exemplar_level}, respectively.

\subsection{Global BOP} 
\label{subsec_global_training_flow}

In MCIL, the classification model is incrementally trained in each phase on the union of new class data and old class \emph{mnemonics} exemplars.
In turn, based on this model, the new class \emph{mnemonics} exemplars (i.e., the parameters of the exemplars) are trained before omitting new class data.
In this way, the optimality of model derives a constrain to optimizing the exemplars, and vise versa.
We propose to formulate this relationship with a global BOP in which each phase uses the optimal model to optimize exemplars, and vice versa.

Specifically, in the $i$-th phase,
our MCIL system aims to learn a model $\Theta_i$ to approximate the \emph{ideal} one named $\Theta^*_i$ which minimizes the classification loss $\mathcal{L}_c$ on both $D_i$ and $D_{0:i-1}$, i.e.,
\begin{equation}\label{eq_main_problem_incremental}
      \Theta_i^* = \argmin_{\Theta_i} \mathcal{L}_c(\Theta_i; D_{0:i-1}\cup D_i).
\end{equation}
Since $D_{0:i-1}$ was omitted (i.e., not accessible) and only $\mathcal{E}_{0:i-1}$ is stored in memory, we approximate $\mathcal{E}_{0:i-1}$ towards the optimal replacement of $D_{0:i-1}$ as much as possible.
We formulate this with the global BOP, where ``global'' means operating through all phases, as follows,
\begin{subequations}
\label{eq_global_bilevel_program}
\begin{align}\label{eq_global_bilevel_upper}
    &\min_{\Theta_i} \mathcal{L}_{c}(\Theta_i; \mathcal{E}^*_{0:i-1}\cup D_i) \\
\label{eq_global_bilevel_lower}
    &\ \text{s.t.} \ \mathcal{E}^*_{0:i-1} = \argmin_{\mathcal{E}_{0:i-1}} \mathcal{L}_{c}\big(\Theta_{i-1}(\mathcal{E}_{0:i-1}); \mathcal{E}_{0:i-2}\cup D_{i-1}\big), %\nonumber
\end{align}
\end{subequations}
where $\Theta_{i-1}(\mathcal{E}_{0:i-1})$ denotes that $\Theta_{i-1}$ was fine-tuned on $\mathcal{E}_{0:i-1}$ to reduce the bias caused by the imbalanced sample numbers between new class data $D_{i-1}$ and old exemplars $\mathcal{E}_{0:i-2}$, in the $i-1$-th phase. Please refer to the last paragraph in Section~\ref{subsect_exemplar_level} for more details.
In the following paper, Problem~\ref{eq_global_bilevel_upper} and Problem~\ref{eq_global_bilevel_lower} are called \emph{model-level} and \emph{exemplar-level} problems, respectively.

\subsection{Model-level problem}
\label{subsec_model_level}

As illustrated in Figure~\ref{fig_framework}, in the $i$-th phase, we first solve the \emph{model-level} problem with the \emph{mnemonics} exemplars $\mathcal{E}_{0:i-1}$ as part of the input and previous $\Theta_{i-1}$ as the model initialization. 
According to Problem~\ref{eq_global_bilevel_program}, the objective function can be expressed as
\begin{equation}\label{eq_full_loss}
    \mathcal{L}_{\text{all}} = \lambda\mathcal{L}_{c}(\Theta_i; \mathcal{E}_{0:i-1}\cup D_i) \\
    +(1-\lambda)\mathcal{L}_{d}(\Theta_i; \Theta_{i-1}; \mathcal{E}_{0:i-1}\cup D_i),
\end{equation}
where $\lambda$ is a scalar manually set to balance between $\mathcal{L}_{d}$ and $\mathcal{L}_{c}$ (introduced in Section~\ref{sec_preli}).
Let $\alpha_1$ be the learning rate, $\Theta_i$ is updated with gradient descent as follows,
\begin{equation}\label{eq_model_theta_update}
    \Theta_i \gets \Theta_i - \alpha_1 \nabla_{\Theta}\mathcal{L}_{\text{all}}.
\end{equation}
Then, $\Theta_i$ will be used to train the parameters of the \emph{mnemonics} exemplars, i.e., to solve the \emph{exemplar-level} problem in Section~\ref{subsect_exemplar_level}.

\subsection{Exemplar-level problem}
\label{subsect_exemplar_level}

Typically, the number of exemplars $\mathcal{E}_i$ is set to be greatly smaller than that of the original data $D_i$.
Existing methods~\cite{rebuffi2017icarl, hou2019learning, Wu2019LargeScale, Castro18EndToEnd} are always based on the assumption that the models trained on the few exemplars also minimize its loss on the original data. However, there is no guarantee particularly when these exemplars are heuristically chosen.
In contrast, our approach explicitly aims to ensure a feasible approximation of that assumption, thanks to the differentiability of our \emph{mnemonics} exemplars.

To achieve this, we train a temporary model $\Theta'_i$ on $\mathcal{E}_i$ to maximize the prediction on $D_i$, for which we use $D_i$ to compute a validation loss to penalize this temporary training with respect to the parameters of $\mathcal{E}_i$.
The entire problem is thus formulated in a local BOP, where ``local'' means within a single phase, as 
\begin{subequations}\label{eq_omi_bilevel_program}
\begin{align}\label{eq_omi_bilevel_upper}
  &\min_{\mathcal{E}_i} \mathcal{L}_{c}\big(\Theta'_i(\mathcal{E}_i); D_i\big) \\
\label{eq_omi_bilevel_lower}
  &\ \text{s.t.} \ \Theta'_i(\mathcal{E}_i) = \argmin_{\Theta_i} \mathcal{L}_{c}(\Theta_i; \mathcal{E}_i).
\end{align}
\end{subequations}
We name the temporary training in Problem~\ref{eq_omi_bilevel_lower} as \emph{base-level} optimization and the validation in Problem~\ref{eq_omi_bilevel_upper} as \emph{meta-level} optimization, similar to the naming in meta-learning applied to tackling few-shot tasks~\cite{finn2017model}.

\myparagraph{Training $\mathcal{E}_i$.}
The training flow is detailed in Figure~\ref{fig_framework_details}(b) with the data split on the left of Figure~\ref{fig_framework_details}(a).
First, the image-size parameters of $\mathcal{E}_i$ are initialized by a random sample subset $S$ of $D_i$.
Second, we initialize a temporary model $\Theta'_i$ using $\Theta_i$ and train $\Theta'_i$ on $\mathcal{E}_i$ (denoted uniformly as $\mathcal{E}$ in~\ref{fig_framework_details}(b)), for a few iterations by gradient descent:
\begin{equation}\label{eq_shadow_theta_update}
    \Theta'_i \gets \Theta'_i - \alpha_2 \nabla_{\Theta'}\mathcal{L}_c(\Theta'_i; \mathcal{E}_i),
\end{equation}
where $\alpha_2$ is the learning rate of fine-tuning temporary models.
Finally, as the $\Theta_{i}'$ and $\mathcal{E}_i$ are both differentiable, we are able to compute the loss of $\Theta'_i$ on $D_i$, and back-propagate this validation loss to optimize $\mathcal{E}_i$,
\begin{equation}\label{eq_update_ei_sgd}
    \mathcal{E}_i \gets  \mathcal{E}_i - \beta_1 \nabla_{\mathcal{E}}\mathcal{L}_{c}\big(\Theta'_i(\mathcal{E}_i); D_i\big),
\end{equation}
where $\beta_1$ is the learning rate.
In this step, we basically need to back-propagate the validation gradients till the input layer, through unrolling all training gradients of $\Theta'_i$.
This operation involves a gradient through a gradient. Computationally, it requires an additional backward pass through $\mathcal{L}_c(\Theta'_i;\mathcal{E}_i)$ to compute Hessian-vector products, which is supported by standard numerical computation libraries such as TensorFlow~\cite{AbadiABBCCCDDDG16} and PyTorch~\cite{steiner2019pytorch}.

\myparagraph{Adjusting $\mathcal{E}_{0:i-1}$.}
The \emph{mnemonics} exemplars of a previous class were trained when this class occurred. It is desirable to adjust them to the changing data distribution online.
However, old class data $D_{0:i-1}$ are not accessible, so it is not feasible to directly apply Eq.~\ref{eq_update_ei_sgd}. 
Instead, we propose to split $\mathcal{E}_{0:i-1}$ into two subsets and subject to $\mathcal{E}_{0:i-1}=\mathcal{E}^{\textbf{A}}_{0:i-1}\cup\mathcal{E}^{\textbf{B}}_{0:i-1}$.
We use one of them, e.g. $\mathcal{E}^{\textbf{B}}_{0:i-1}$, as the validation set (i.e., a replacement of $D_{0:i-1}$) to optimize the other one, e.g., $\mathcal{E}^{\textbf{A}}_{0:i-1}$, as shown on the right of Figure~\ref{fig_framework_details}(a). Alternating the input and target data in Figure~\ref{fig_framework_details}(b), we adjust all old exemplars in two steps:
\begin{subequations}\label{eq_update_e_sgd_AB_total}
\begin{equation}\label{eq_update_e_sgdAB}
   \mathcal{E}^{\textbf{A}}_{0:i-1} \gets  \mathcal{E}^{\textbf{A}}_{0:i-1} - \beta_2 \nabla_{\mathcal{E}^{\textbf{A}}}\mathcal{L}_{c}\big(\Theta'_i(\mathcal{E}^{\textbf{A}}_{0:i-1}); \mathcal{E}^{\textbf{B}}_{0:i-1}\big),
\end{equation}
\begin{equation}\label{eq_update_e_sgdBA}
   \mathcal{E}^{\textbf{B}}_{0:i-1} \gets  \mathcal{E}^{\textbf{B}}_{0:i-1} - \beta_2 \nabla_{\mathcal{E}^{\textbf{B}}}\mathcal{L}_{c}\big(\Theta'_i(\mathcal{E}^{\textbf{B}}_{0:i-1}); \mathcal{E}^{\textbf{A}}_{0:i-1}\big),
\end{equation}
\end{subequations}
where $\beta_2$ is the learning rate. $\Theta'_i(\mathcal{E}^{\textbf{B}}_{0:i-1})$ and $\Theta'_i(\mathcal{E}^{\textbf{A}}_{0:i-1})$ are trained by replacing $\mathcal{E}_i$ in Eq.~\ref{eq_shadow_theta_update} with $\mathcal{E}^{\textbf{B}}_{0:i-1}$ and $\mathcal{E}^{\textbf{A}}_{0:i-1}$, respectively. We denote the adjusted exemplars as $\tilde{\mathcal{E}}_{0:i-1}$. Note that we can also split $\mathcal{E}_{0:i-1}$ into more than $2$ subsets, and optimize each subset using its complement as the validation data, following the same strategy in Eq.~\ref{eq_update_e_sgd_AB_total}.

\myparagraph{Fine-tuning models on only exemplars.}
The model $\Theta_i$ has been trained on $D_i\cup\mathcal{E}_{0:i-1}$, and may suffer from the classification bias caused by the imbalanced sample numbers, e.g., $1000$ \emph{versus} $20$, between the classes in $D_i$ and $\mathcal{E}_{0:i-1}$.
To alleviate this bias, we propose to fine-tune $\Theta_i$ on $\mathcal{E}_i\cup\tilde{\mathcal{E}}_{0:i-1}$ in which each class has exactly the same number of samples (exemplars).

\section{Weight transfer operations} 
\label{suppsec_transfer_weights}

We deploy weight transfer operations~\cite{sun2019meta, FiLM2018} to train the weight scaling and shifting parameters (named $\mathbf{T}_i$) in the $i$-th phase which specifically transfer the network weights $\Theta_{i-1}$ to $\Theta_{i}$.
The aim is to preserve the structural knowledge of $\Theta_{i-1}$ when learning $\Theta_{i}$ on new class data. 

Specifically, we assume the $q$-th layer of $\Theta_{i-1}$ contains $R$ neurons, so we have $R$ neuron weights and biases as $\{W_{q,r}, b_{q,r}\}_{r=1}^R$. For conciseness, we denote them as $W_{q}$ and $b_{q}$.
For $W_{q}$, we learn $R$ scaling parameters denoted as $\mathcal{T}^W_{q}$, and for $b_{q}$, we learn $R$ shifting parameters denoted as $\mathcal{T}^b_{q}$.
Let $X_{q-1}$ and $X_q$ be the input and output (feature maps) of the $q$-th layer. 
We apply $\mathcal{T}^W_q$ and $\mathcal{T}^b_q$ to $W_q$ and $b_q$ as,
\begin{equation}\label{eq_T_operation_layer}
     X_q =(W_q\odot\mathcal{T}^W_{q}) X_{q-1} + (b_q + \mathcal{T}^b_{q}),
\end{equation}
where $\odot$ donates the element-wise multiplication. Assuming there are $Q$ layers in total, the scaling and shifting parameters are denoted as
$\mathbf{T}_i=\{\mathcal{T}^W_{q}, \mathcal{T}^b_{q}\}_{q=1}^Q$.

Therefore, to learn the MCIL model $\Theta_{i}$, we use the indirect way of training $\mathbf{T}_i$ (instead of the direct way of training $\Theta_{i}$) on $D_i \cup \mathcal{E}_{0:i-1}$ and keeping $\Theta_{i-1}$ fixed. 
During the training, both classification loss and distillation loss~\cite{Li18LWF, rebuffi2017icarl} are used. Let $\odot_{L}$ donate the function of applying $\mathbf{T}_i$ to $\Theta_{i-1}$ by layers (Eq.~\ref{eq_full_loss_supp}). 
The objective function Eq.~\textcolor{red}{4} in the main paper can be rewritten as:
\begin{equation}\label{eq_full_loss_supp}
    \begin{split}
    \mathcal{L}_{\text{all}} =& \lambda\mathcal{L}_{c}(\mathbf{T}_i\odot_{L}\Theta_{i-1}; \mathcal{E}_{0:i-1}\cup D_i)  \\
    &+(1-\lambda)\mathcal{L}_{d}(\mathbf{T}_i\odot_{L}\Theta_{i-1}; \Theta_{i-1}; \mathcal{E}_{0:i-1}\cup D_i),
    \end{split}
\end{equation}
where $\lambda$ is a scalar manually set to balance two loss terms.
Let $\alpha_1$ be the learning rate, $\mathbf{T}_i$ is updated with gradient descent as follows,
\begin{equation}\label{eq_model_theta_update_supp}
    \mathbf{T}_i \gets \mathbf{T}_i - \alpha_1 \nabla_{\mathbf{T}}\mathcal{L}_{\text{all}}.
\end{equation}
After the learning of $\mathbf{T}_i$, we compute $\Theta_i$ as follows:
\begin{equation}\label{eq_T_total}
    \Theta_i \gets \mathbf{T}_i \odot_{L} \Theta_{i-1}.
\end{equation}

\begin{algorithm}
\caption{Mnemonics Training}
\label{alg_overall}
\SetAlgoLined
\SetKwInput{KwData}{Input}
\SetKwInput{KwResult}{Output}
 \KwData{Data flow $\{D_i\}_{i=0}^N$.}
 \KwResult{MCIL models $\{\Theta_i\}_{i=0}^N$, and \emph{mnemonics} exemplars $\{\mathcal{E}_i\}_{i=0}^N$.}
\For{$i$ \emph{\textbf{in}} $\{0,1,...,N\}$ }
{
Get $D_i$\;
\uIf{$i = 0$}{
Randomly initialize $\Theta_0$ and train it on $D_0$\;
}
\Else{
Get $\mathcal{E}_{0:i-1}$ from memory\;
Initialize $\Theta_i$ with $\Theta_{i-1}$\;
Train $\Theta_i$ on $\mathcal{E}_{0:i-1}\cup D_i$ by Eq.~\ref{eq_model_theta_update}\;
}
Sample $S$ from $D_i$ to initialize $\mathcal{E}_i$\;
Train $\mathcal{E}_i$ using $\Theta_i$ by Eq.~\ref{eq_update_ei_sgd}\;

\While{$i\geq1$ }{
Split $\mathcal{E}_{0:i-1}$ into subsets $\mathcal{E}^{\textbf{A}}_{0:i-1}$ and $\mathcal{E}^{\textbf{B}}_{0:i-1}$ \;
Optimize $\mathcal{E}^{\textbf{A}}_{0:i-1}$ and $\mathcal{E}^{\textbf{B}}_{0:i-1}$ by Eq.~\ref{eq_update_e_sgd_AB_total}\;
Get the adjusted old exemplars $\tilde{\mathcal{E}}_{0:i-1}$
}

(Optional) delete part of the exemplars in $\tilde{\mathcal{E}}_{0:i-1}$\;
Finetune $\Theta_i$ on $\mathcal{E}_i\cup\tilde{\mathcal{E}}_{0:i-1}$\;
Run test and record the results\;
Update $\mathcal{E}_{0:i}\gets\mathcal{E}_i\cup\tilde{\mathcal{E}}_{0:i-1}$ in memory.
}
\end{algorithm}

\section{Algorithm}
\label{subsec_classification_model_details}

In Algorithm~\ref{alg_overall}, we summarize the overall process of the proposed \emph{mnemonics} training. 
Step~1-16 show the alternative learning of classification models and \emph{mnemonics} exemplars, corresponding to Sections~\redt{4.1}-\redt{4.3}.
Specifically in each phase, Step~8 executes the \emph{model-level} training, while Step~11 and~14 are the \emph{exemplar-level}.
Step~17 is optional due to different MCIL settings regarding the memory budget. 
We conduct experiments in two settings: (1) each class has a fixed number (e.g., $20$) of exemplars, and (2) the system consistently keeps a fixed memory budget in all phases, therefore, the system in earlier phases can store more exemplars per class and needs to discard old exemplars in later phases gradually.
Step~18 fine-tunes the model on adjusted and balanced examples. It is helpful to reduce the previous model bias (Step~8) caused by the imbalance samples numbers between new class data $D_i$ and old exemplars $\mathcal{E}_{0:i-1}$.
Step~19 is to evaluate the learned model $\Theta_i$ in the current phase, and the average over all phases will be reported as the final evaluation.
Step~20 updates the memory to include new exemplars.

\section{Experiments}
\label{sec_experiments}
\begin{figure*}
\newcommand{\newincludegraphics}[1]{\includegraphics[height=1.26in]{#1}}
\centering
\vspace{0.2cm}
\includegraphics[height=0.16in]{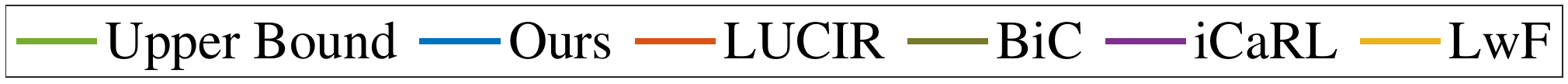}
\vspace{0.4cm}
\subfigure[CIFAR-100 ($100$ classes). In the $0$-th phase, $\Theta_0$ is trained on $50$ classes, the remaining  classes are given evenly in the subsequent phases.]{
%\newincludegraphics{files/cifar100_5.pdf}
\newincludegraphics{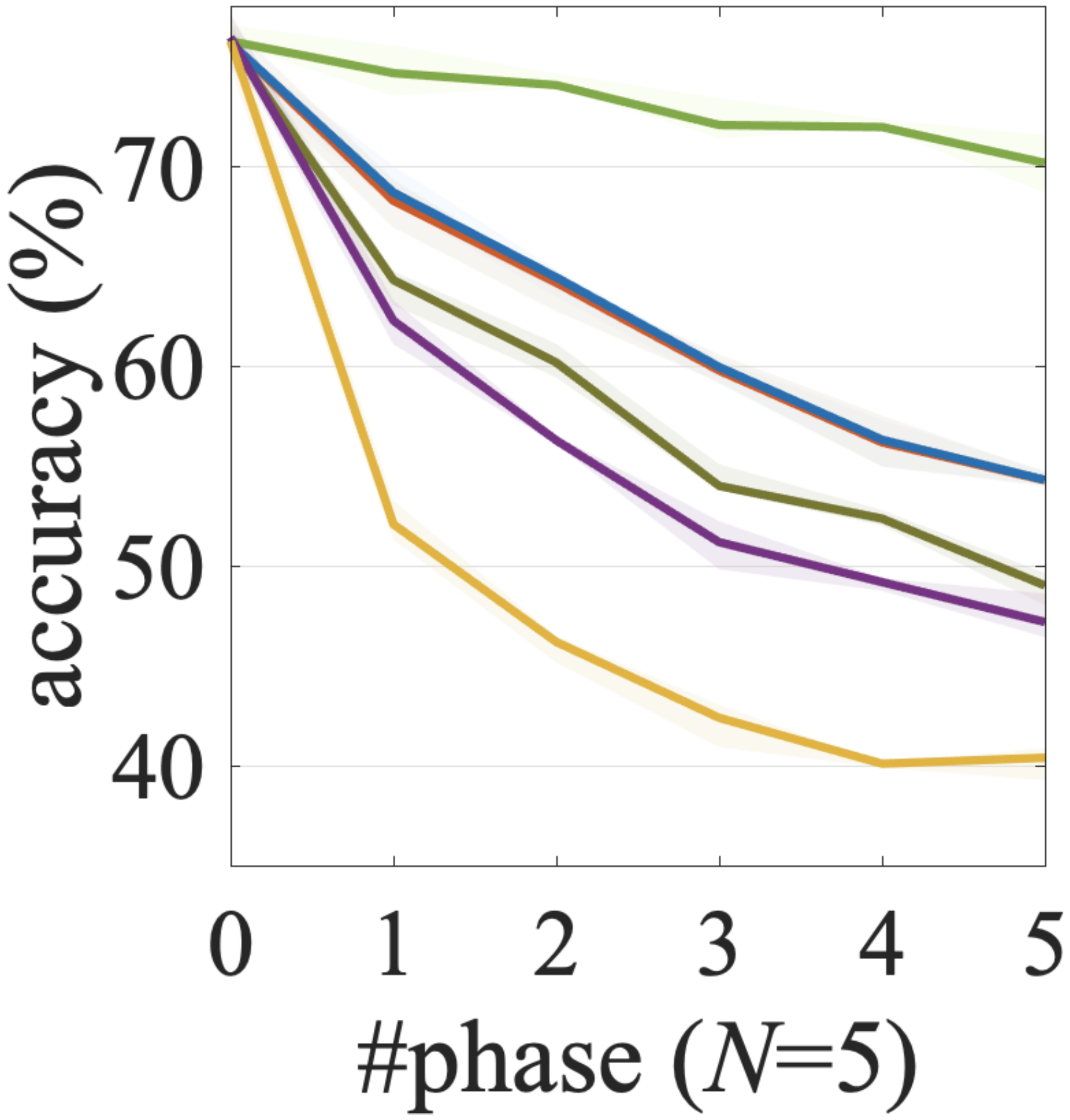}

\hspace{1mm}
\newincludegraphics{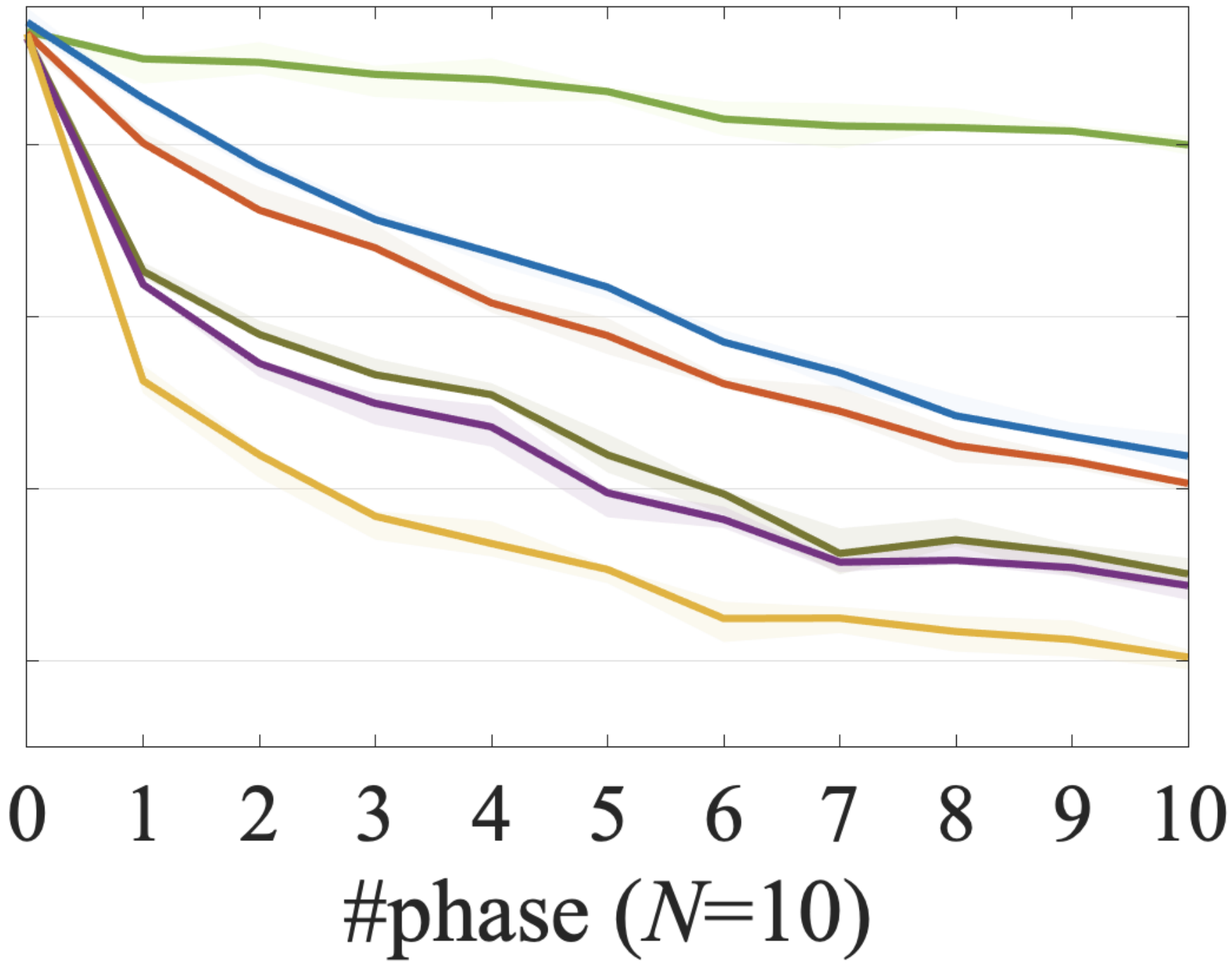}
\hspace{1mm}
\newincludegraphics{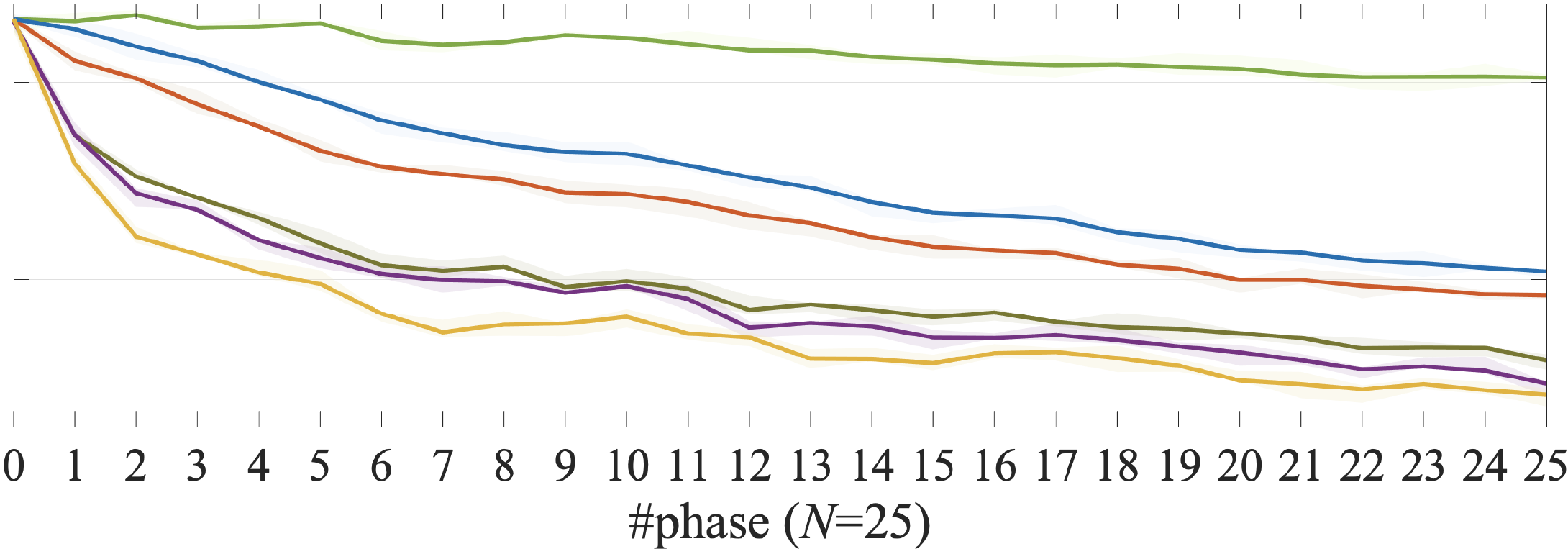}
}

\vspace{0.2cm}
\subfigure[ImageNet-Subset ($100$ classes). In the $0$-th phase, $\Theta_0$ is trained on $50$ classes, the remaining classes are given evenly in the subsequent phases.]{
\newincludegraphics{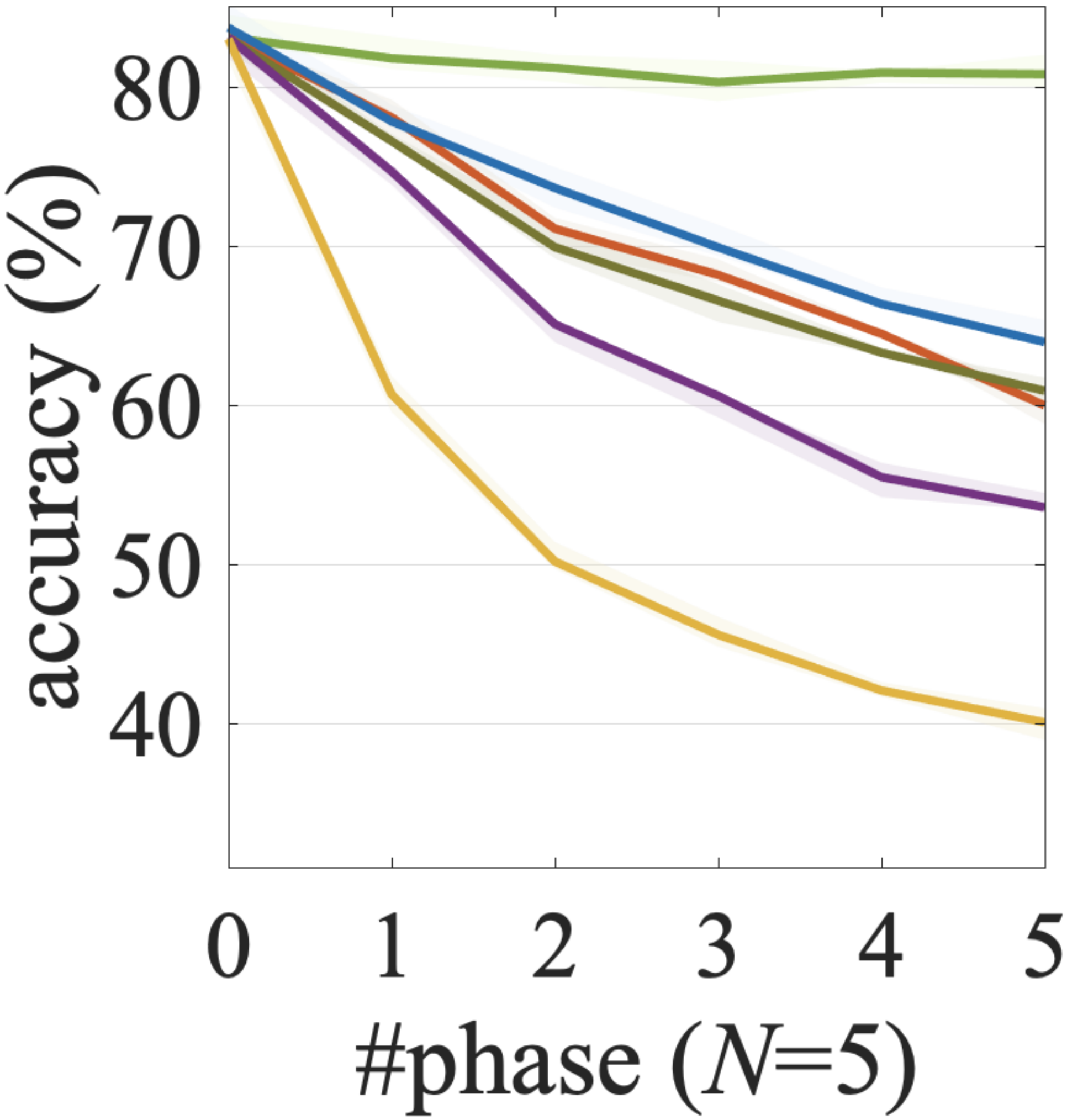}

\hspace{1mm}
\newincludegraphics{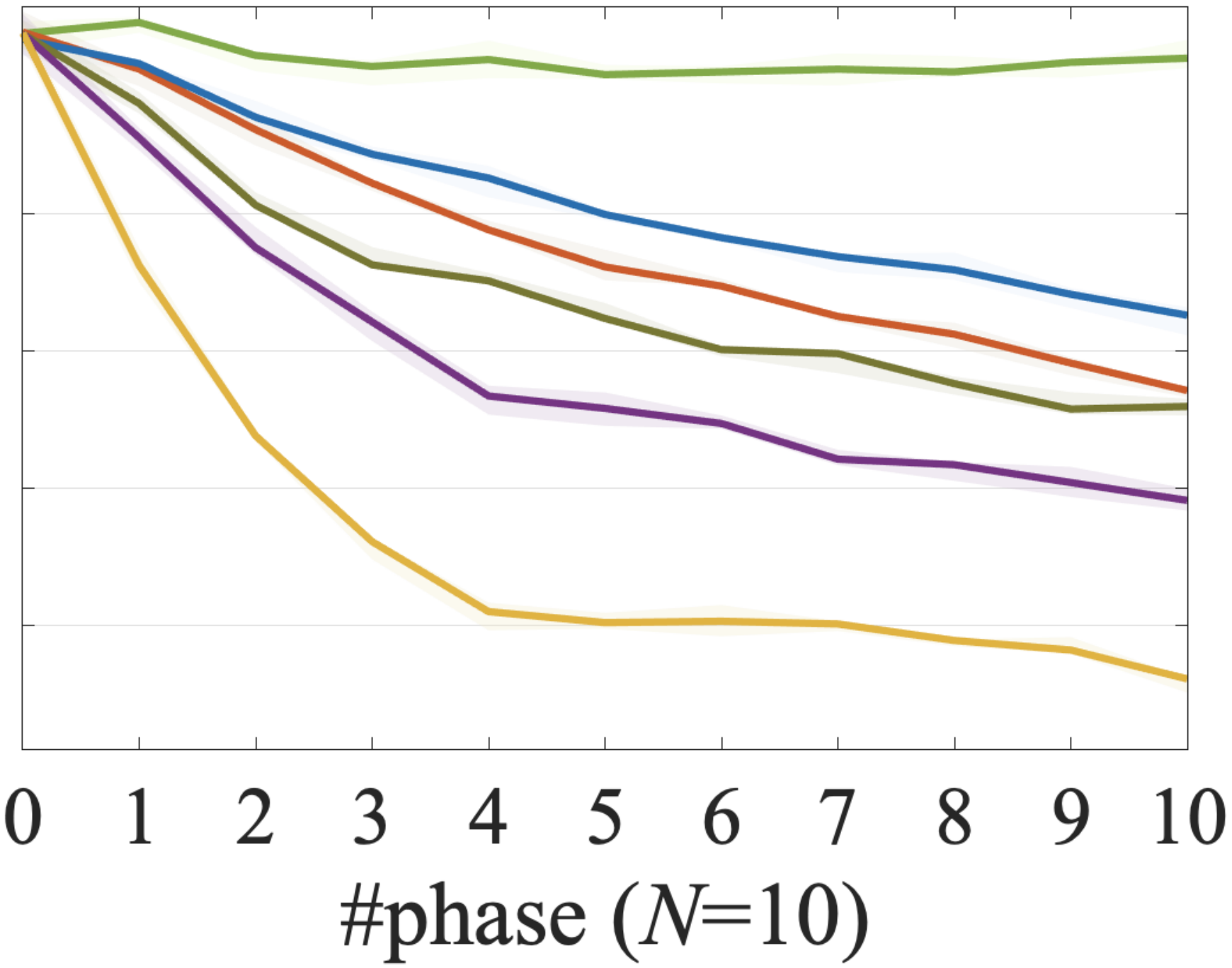}

\hspace{1mm}
\newincludegraphics{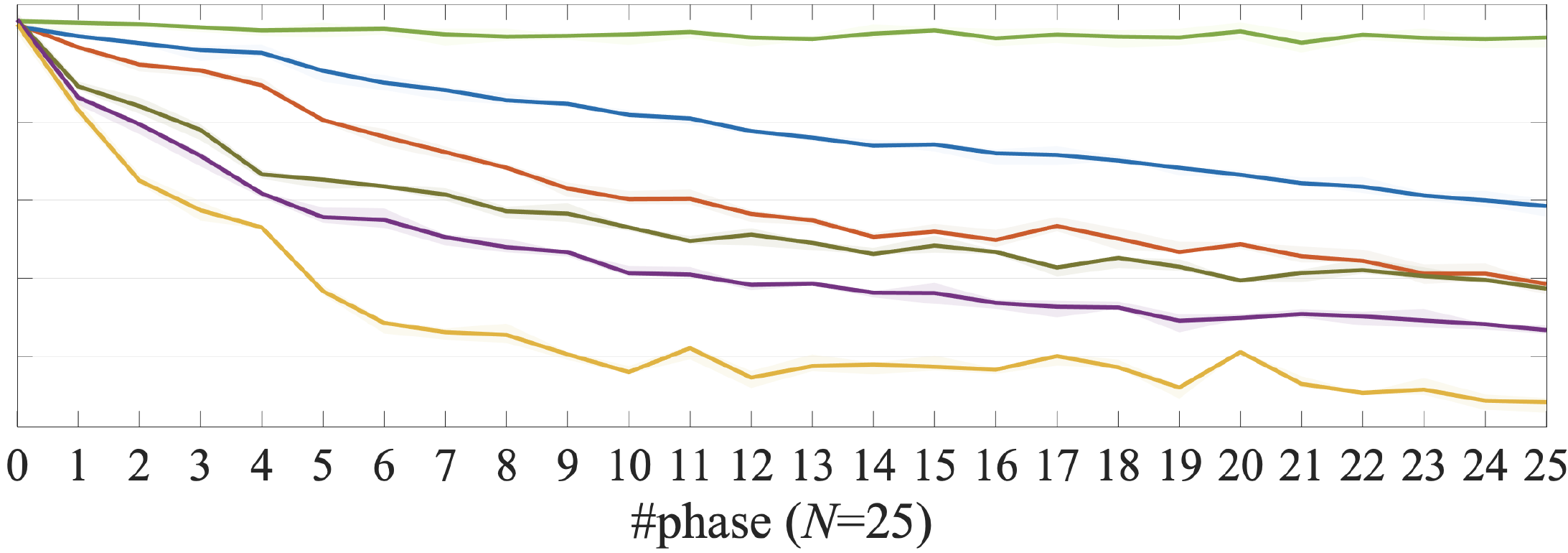}
}

\vspace{0.5cm}
\subfigure[ImageNet ($1000$ classes). In the $0$-th phase, $\Theta_0$ on is trained on $500$ classes, the remaining classes are given evenly in the subsequent phases.]{
\newincludegraphics{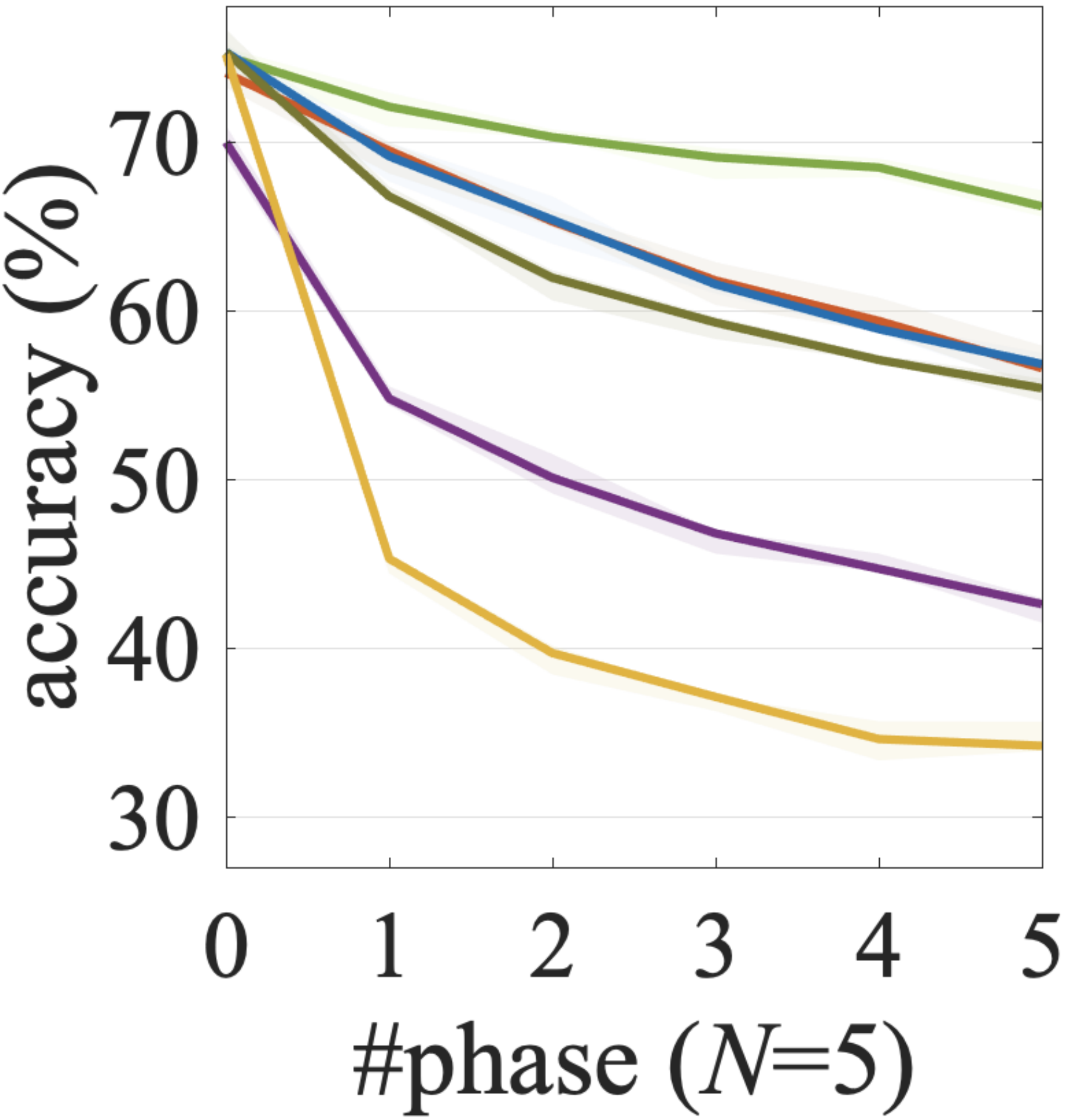}

\hspace{1mm}

\newincludegraphics{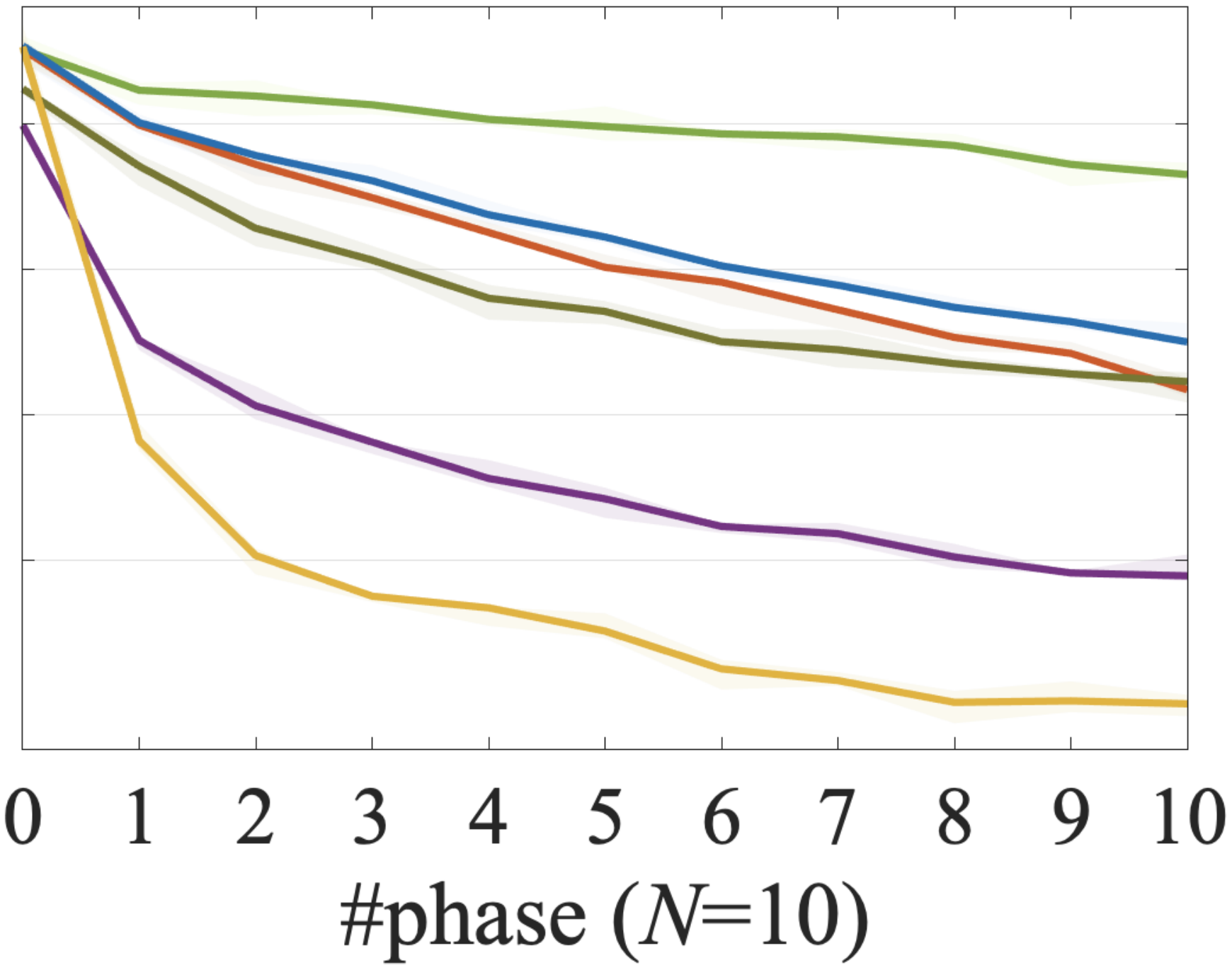}

\hspace{1mm}

\newincludegraphics{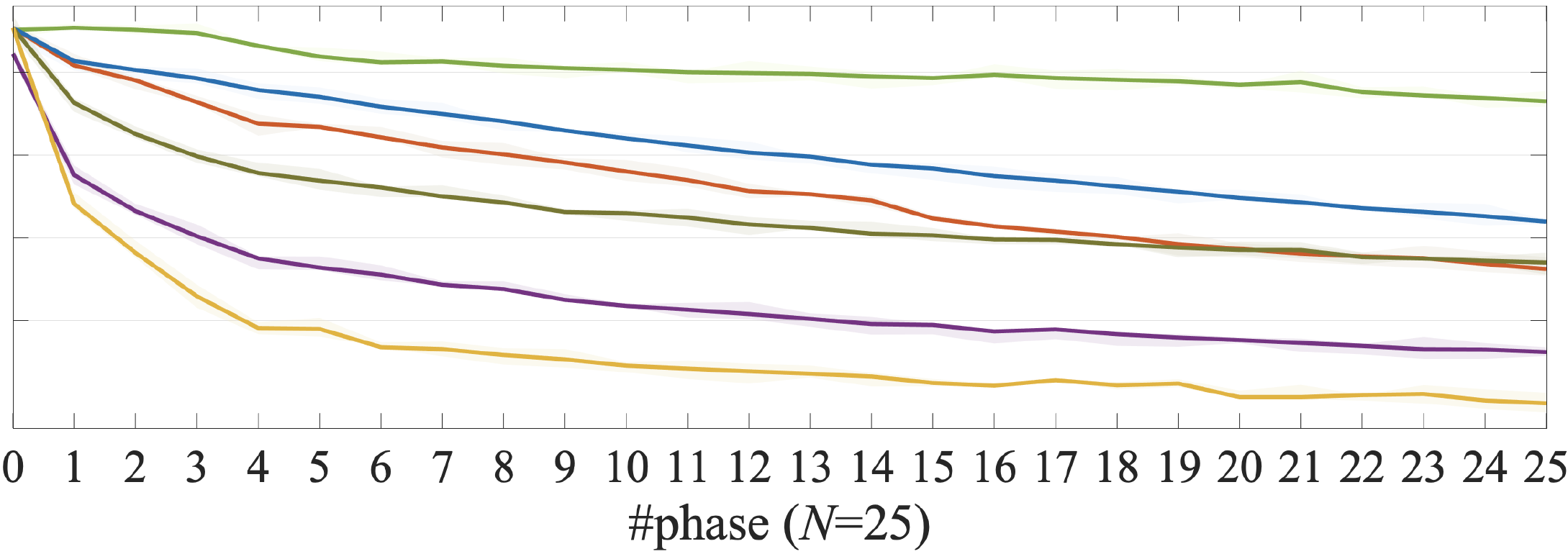}
}
\vspace{0.2cm}
\cotronlcaptionvsapce
\caption{Phase-wise accuracies ($\%$).
Light-color ribbons are visualized to show the $95\%$ confidence intervals. 
Comparing methods: Upper Bound (the results of joint training with all previous data accessible in each phase); LUCIR (2019)~\cite{hou2019learning}; BiC (2019)~\cite{Wu2019LargeScale}; iCaRL (2017)~\cite{rebuffi2017icarl}; and LwF (2016)~\cite{Li18LWF}. 
We show Ours results using ``LUCIR \emph{w/} ours''. Please refer to the average accuracy of each curve in Table~\ref{table_sota_new}. 
}
\label{figure_acc_plots}
\cotronlvsapce
%\vspace{0.3cm}
\end{figure*}

\begin{table*}[htp]
  \vspace{-0.3cm}
  \small
  \centering
  \begin{tabular}{llccccccccccc}
  \toprule
  \multirow{2.5}{*}{Metric} & \multirow{2.5}{*}{Method} & \multicolumn{3}{c}{\emph{CIFAR-100}} && \multicolumn{3}{c}{\emph{ImageNet-Subset}} && \multicolumn{3}{c}{\emph{ImageNet}}\\
  \cmidrule{3-5} \cmidrule{7-9} \cmidrule{11-13}
  & & $N$=5 & 10  & 25 && 5 & 10 & 25 && 5 & 10 & 25 \\
   \midrule
   & LwF${}^{\diamond}$ (2016)~\cite{Li18LWF} & 49.59 & 46.98 & 45.51 && 53.62 & 47.64 & 44.32 && 44.35 & 38.90 & 36.87\\
   & LwF \emph{w/} ours & \updatedredt{54.21} & \updatedredt{52.72} & \updatedredt{51.59} && \updatedredt{60.94} & \updatedredt{59.25} & 59.71 && 52.70 & 50.37 & 50.79 \\
   \cmidrule{2-13}
   & iCaRL (2017)~\cite{rebuffi2017icarl} & 57.12 & 52.66 & 48.22 && 65.44 & 59.88 & 52.97 && 51.50 & 46.89 & 43.14  \\
   \emph{Average acc.} (\%) $\uparrow$ & iCaRL \emph{w/} ours & \updatedredt{60.00} & \updatedredt{57.37} & \updatedredt{54.13} && \updatedredt{72.34} & \updatedredt{70.50} & 67.12 && 60.61 & 58.62 & 53.46  \\
   \cmidrule{2-13}
    \multirow{1}{*}{$\bar{\mathcal{A}}=\frac{1}{N+1}\sum_{i=0}^{N}\mathcal{A}_i$} & BiC (2019)~\cite{Wu2019LargeScale} & 59.36 & 54.20 & 50.00 && 70.07 & 64.96 & 57.73 && 62.65 & 58.72 & 53.47 \\
   & BiC \emph{w/} ours & 60.67 & 58.11 & 55.51 && 71.92 & 70.73 & 69.22 && \textbf{64.63} & 62.71 & 60.20 \\
   \cmidrule{2-13}
   & LUCIR (2019)~\cite{hou2019learning}  & 63.17 & 60.14 & 57.54 && 70.84 & 68.32 & 61.44 && 64.45 & 61.57 & 56.56 \\
   & LUCIR \emph{w/} ours & {\textbf{63.34}} & {\textbf{62.28}} &  {\textbf{60.96}} &&  {\textbf{72.58}} &  {\textbf{71.37}} &  {\textbf{69.74}} &&  {\redt{64.54}} &  {\textbf{63.01}} &  {\textbf{61.00}} \\
  \midrule
   & LwF${}^{\diamond}$ (2016)~\cite{Li18LWF} & 43.36 & 43.58 & 41.66 && 55.32 & 57.00 & 55.12 && 48.70 & 47.94 & 49.84\\
   & LwF \emph{w/} ours & \updatedredt{40.00} & \updatedredt{36.50} & \updatedredt{34.26} && \updatedredt{41.07} & \updatedredt{39.76} & 39.99 && 37.46 & 38.42 & 37.95 \\
   \cmidrule{2-13}
   & iCaRL (2017)~\cite{rebuffi2017icarl} & 31.88 & 34.10 & 36.48 && 43.40 & 45.84 & 47.60 && 26.03 & 33.76 & 38.80\\
   \emph{Forgetting rate} (\%) $\downarrow$ & iCaRL \emph{w/} ours & \updatedredt{25.94} & \updatedredt{26.92} & \updatedredt{28.92} && \updatedredt{20.96} & \updatedredt{24.12} & 29.32 && 20.26 & 24.04 & 17.49  \\
   \cmidrule{2-13} 
   \multirow{1}{*}{$\mathcal{F}=\mathcal{A}_{N}^Z-\mathcal{A}^Z_0$}  & BiC (2019)~\cite{Wu2019LargeScale} & 31.42 & 32.50 & 34.60 && 27.04 & 31.04 & 37.88 && 25.06 & 28.34 & 33.17\\
   & BiC \emph{w/} ours & 22.42 & 24.50 & 25.52 && 18.43 & 19.20 & 21.43 && 18.32 & 19.72 & 20.50\\
   \cmidrule{2-13}   
   & LUCIR (2019)~\cite{hou2019learning}  & 18.70 & 21.34 & 26.46 && 31.88 & 33.48 & 35.40 && 24.08 & 27.29 & 30.30\\
   & LUCIR \emph{w/} ours &  {\textbf{10.91}} &  {\textbf{13.38}} &  {\textbf{19.80}} &&  {\textbf{17.40}} &  {\textbf{17.08}} &  {\textbf{20.83}} &&  {\textbf{13.85}} &  {\textbf{15.82}} &  {\textbf{19.17}}\\
  \bottomrule

    \multicolumn{13}{l}{${}^{\diamond}$ Using \emph{herding} exemplars as~\cite{hou2019learning,rebuffi2017icarl,Wu2019LargeScale} for fair comparison.}\\
\end{tabular}
  \cotronlcaptionvsapce
  \caption{
  Average accuracies $\bar{\mathcal{A}}$ (\%) and forgetting rates $\mathcal{F}$ (\%) for the state-of-the-art~\cite{hou2019learning} and other baseline architectures~\cite{Li18LWF,rebuffi2017icarl,Wu2019LargeScale} with and without our \emph{mnemonics} training approach as a plug-in module.  
  Let $D_i^{\text{test}}$ be the test data corresponding to $D_i$ in the $i$-th phase. 
  $\mathcal{A}_i$ denotes the average accuracy of $D_{0:i}^{\text{test}}$ by $\Theta_{i}$. $\mathcal{A}_i^Z$ is the average accuracy of $D_0^{\text{test}}$ by $\Theta_{i}$ in the $i$-th phase.
  Note that the weight transfer operations are applied in ``\emph{w/} ours'' methods.
  }
  \label{table_sota_new}
  \cotronlvsapce
  \vspace{-0.25cm}
\end{table*}

We evaluate the proposed \emph{mnemonics} training approach
on two popular datasets (CIFAR-100~\cite{krizhevsky2009learning} and ImageNet~\cite{russakovsky2015imagenet}) for four different baseline architectures~\cite{Li18LWF,rebuffi2017icarl,Wu2019LargeScale,hou2019learning}, and achieve consistent improvements. Below we describe the datasets and implementation details (Section~\ref{subsec_datasets}), followed by results and analyses (Section~\ref{subsec_results}), including comparisons to the state-of-the-art, ablation studies and visualization results.

\subsection{Datasets and implementation details}
\label{subsec_datasets}

\myparagraph{Datasets.}
We conduct MCIL experiments on two datasets, CIFAR-100~\cite{krizhevsky2009learning} and ImageNet~\cite{russakovsky2015imagenet}, which are widely used in related works~\cite{rebuffi2017icarl,Castro18EndToEnd,Wu2019LargeScale,hou2019learning}. 
\textbf{CIFAR-100}~\cite{krizhevsky2009learning} contains $60,000$ samples of $32\times32$ color images from $100$ classes. Each class has $500$ training and $100$ test samples.
\textbf{ImageNet} (ILSVRC 2012)~\cite{russakovsky2015imagenet} contains around $1.3$ million
samples of $224\times224$ color images from $1,000$ classes. Each class has about $1,300$ training and $50$ test samples. 
ImageNet is typically used in two MCIL settings~\cite{hou2019learning,rebuffi2017icarl}: one based on only a subset of $100$ classes and the other based on the entire $1,000$ classes. 
The $100$-class data in \textbf{ImageNet-Subeset} 
are randomly sampled from ImageNet with an identical random seed ($1993$) by NumPy, following~\cite{rebuffi2017icarl,hou2019learning}.

\myparagraph{The architectures of $\Theta$.}
Following the uniform setting~\cite{rebuffi2017icarl,Wu2019LargeScale,hou2019learning}, we use a $32$-layer ResNet~\cite{He_CVPR2016_ResNet} for CIFAR-100 and an $18$-layer ResNet for ImageNet.
We deploy the weight transfer operations~\cite{sun2019meta, FiLM2018} to train the network, rather than using standard weight over-writing. This helps to reduce \emph{forgetting} between adjacent models (i.e., $\Theta_{i-1}$ and $\Theta_i$). please refer to the supplementary document for the detailed formulation of weight transfer.

\myparagraph{The architecture of $\mathcal{E}$.} 
It depends on the size of image and the number of exemplars we need.
On the CIFAR-100, each \emph{mnemonics} exemplar is a $32\times32\times3$ tensor. On the ImageNet, it is a $224\times224\times3$ tensor. 
The number of exemplars is set in two manners~\cite{hou2019learning}. (1) $20$ samples are uniformly used for every class. Therefore, the parameter size of the exemplars per class is equal to tensor$\times 20$. \emph{This setting is used in the main paper}. 
(2) The system keeps a fixed memory budget, e.g. at most $2,000$ exemplars in total, in all phases. It thus saves more exemplars per class in earlier phases and discard old exemplars afterwards.
\emph{Due to page limits, the results in this setting are presented in the supplementary document}. 
In both settings, we have the consistent finding that \emph{mnemonics} training is the most efficient approach, surpassing the state-of-the-art by large margins with little computational or parametrization overheads.

\myparagraph{\emph{Model-level} hyperparameters.}
The SGD optimizer is used to train $\Theta$. 
Momentum and weight decay parameters are set to $0.9$ and $0.0005$, respectively.
In each (i.e. $i$-th) phase, the learning rate $\alpha_1$ is initialized as $0.1$.
On the CIFAR-100 (ImageNet), $\Theta_i$ is trained in $160$ ($90$) epochs for which $\alpha_1$ is reduced to its $\frac{1}{10}$ after $80$ ($30$) and then $120$ ($60$) epochs.
In Eq.~\ref{eq_full_loss}, the scalar $\lambda$ and temperature $\tau$ are set to $0.5$ and $2$, respectively, following~\cite{rebuffi2017icarl,hou2019learning}.

\myparagraph{\emph{Exemplar-level} hyperparameters.}
An SGD optimizer is used to update \emph{mnemonics} exemplars $\mathcal{E}_i$ and adjust $\mathcal{E}_{0:i-1}$ (as in Eq.~\ref{eq_update_ei_sgd} and Eq.~\ref{eq_update_e_sgd_AB_total} respectively) in $50$ epochs. 
In each phase, the learning rates $\beta_1$ and $\beta_2$ are initialized as $0.01$ uniformly 
and reduced to their half after every $10$ epochs. 
Gradient descent is applied to update the temporary model $\Theta'$ in $50$ epochs (as in Eq.~\ref{eq_shadow_theta_update}). The learning rate $\alpha_2$ is set to $0.01$. 
We deploy the same set of hyperparameters for fine-tuning $\Theta_i$ on $\mathcal{E}_i\cup\tilde{\mathcal{E}}_{0:i-1}$.

\myparagraph{Benchmark protocol.}
This work follows the protocol in the most recent work --- LUCIR~\cite{hou2019learning}. We also implement all other methods~\cite{rebuffi2017icarl,Castro18EndToEnd, Wu2019LargeScale} on this protocol for fair comparison.
Given a dataset, the model ($\Theta_0$) is firstly trained on half of the classes. Then, the model ($\Theta_i$) learns the remaining classes evenly in the subsequent phases. Assume an MCIL system has $1$ initial phase and $N$ incremental phases. The total number of incremental phases $N$ is set to be $5$, $10$ or $25$ (for each the setting is called ``$N$-phase'' setting). 
At the end of each individual phase, the learned $\Theta_i$ is evaluated on the test data $D_{0:i}^{\text{test}}$ where ``${0:i}$'' denote all seen classes so far. The average accuracy $\bar{\mathcal{A}}$ (over all phases) is reported as the final evaluation~\cite{rebuffi2017icarl, hou2019learning}.
In addition, we propose a forgetting rate, denoted as $\mathcal{F}$, by calculating the difference between the accuracies of $\Theta_0$ and $\Theta_N$ on the same initial test data $D_{0}^{\text{test}}$. The lower forgetting rate is better.

\subsection{Results and analyses}
\label{subsec_results}

\begin{figure}[t]
\centering
\includegraphics[width=3.2in]{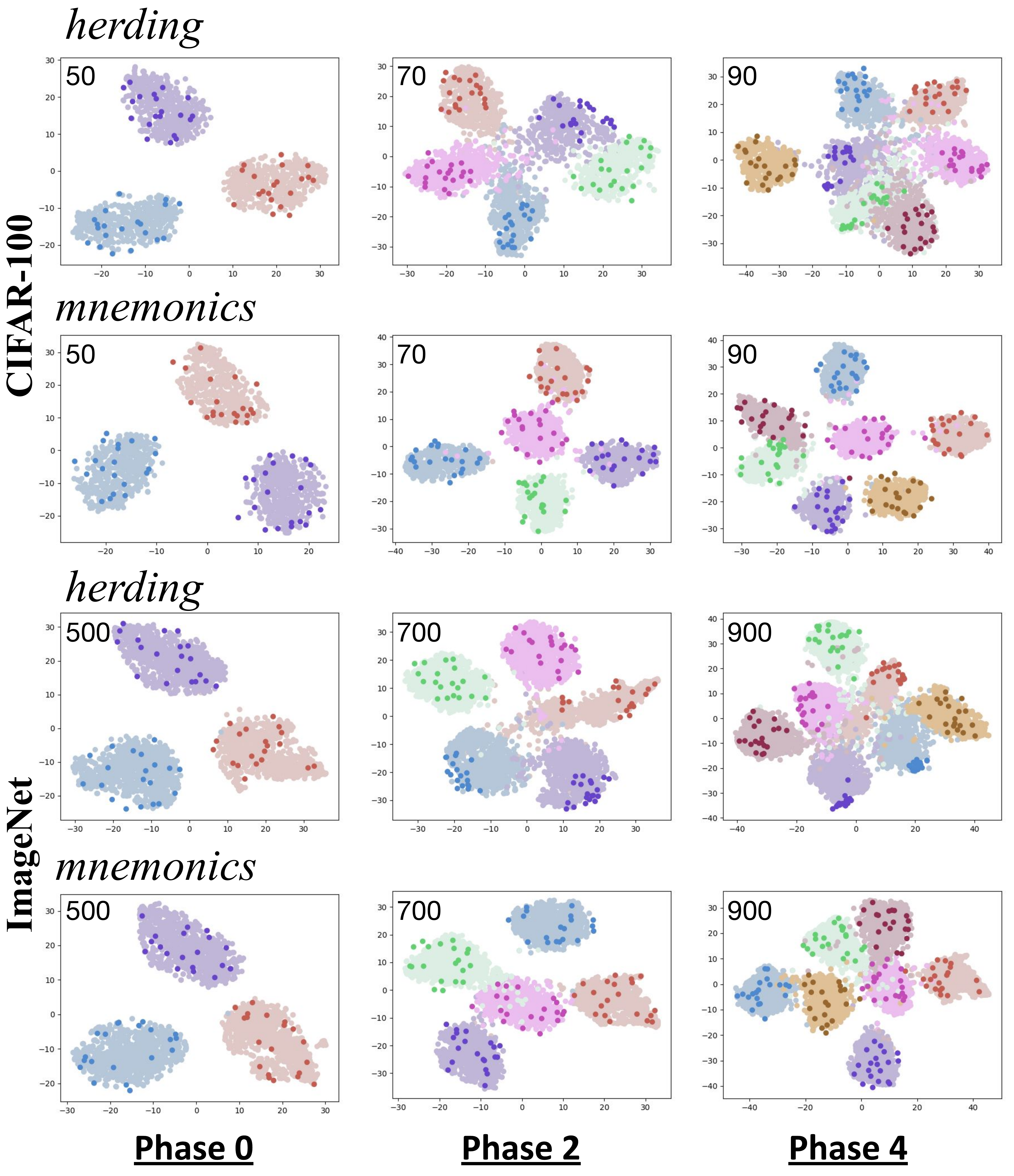}
%\vspace{-0.1cm}
\cotronlcaptionvsapce
\caption{The t-SNE~\cite{maaten2008visualizing}  results of \emph{herding} and our \emph{mnemonics} exemplars on two datasets. $N$=$5$.
In each colored class, deep-color points are exemplars, and light-color ones are original data referring to the real data distribution.
The total number of classes (used in the training) is given in the top-left corner of each sub-figure. For clear visualization,
Phase-$0$ randomly picks $3$ classes from $50$ ($500$) classes on CIFAR-100 (ImageNet). Phase-$2$ and Phase-$4$ increases to $5$ and $7$ classes, respectively.
}
\label{fig_5phase}
\cotronlvsapce
%\vspace{-0.3cm}
\end{figure}

Table~\ref{table_sota_new} shows the comparisons with the state-of-the-art~\cite{hou2019learning} and other baseline architectures~\cite{Li18LWF, rebuffi2017icarl, Wu2019LargeScale}, with and without our \emph{mnemonics} training as a plug-in module. Note that ``without'' in~\cite{Li18LWF, rebuffi2017icarl, Wu2019LargeScale, hou2019learning} means using \emph{herding} exemplars (we add \emph{herding} exemplars to~\cite{Li18LWF} for fair comparison).
Figure~\ref{figure_acc_plots} shows the phase-wise results of our best model, i.e., LUCIR~\cite{hou2019learning} \emph{w/} ours, and those of the baselines.
Table~\ref{table_ablation} demonstrates the ablation study for evaluating two key components: training \emph{mnemonics} exemplars; and adjusting old \emph{mnemonics} exemplars.
Figure~\ref{fig_5phase} visualizes the differences between \emph{herding} and \emph{mnemonics} exemplars in the data space.

\myparagraph{Compared to the state-of-the-art.} 
Table~\ref{table_sota_new} shows that taking our 
\emph{mnemonics} training 
as a plug-in module on the state-of-the-art~\cite{hou2019learning} and other baseline architectures
consistently improves their performance.
In particular, LUCIR~\cite{hou2019learning} \emph{w/} ours achieves the highest average accuracy
and lowest forgetting rate, e.g. respectively $61.00\%$ and $19.17\%$ on the most challenging $25$-phase ImageNet.
The overview on forgetting rates $\mathcal{F}$ reveals that our approach is greatly helpful to reduce forgetting problems for every method. 
For example, LUCIR (\emph{w/} ours) sees its $\mathcal{F}$ reduced to around the third and the half on the $25$-phase CIFAR-100 and ImageNet, respectively.

\myparagraph{Different total phases ($N$ = $5$, $10$, $25$).} Table~\ref{table_sota_new} and Figure~\ref{figure_acc_plots} demonstrate that the boost by our \emph{mnemonics} training becomes larger in more-phase settings, e.g. on ImageNet-Subset, LUCIR \emph{w/} ours gains $1.74\%$ on $5$-phase while $8.30\%$ on $25$-phase.
When checking the ending points of the curves from $N$=$5$ to $N$=$25$ in Figure~\ref{figure_acc_plots}, we find related methods, LUCIR, BiC, iCaRL and LwF, all suffer from performance drop.
The possible reason is that their models get more and more seriously overfitted to \emph{herding} exemplars which are heuristically chosen and fixed.
In contrast, our best model (LUCIR \emph{w/} ours) does not have such problem, thanks for our \emph{mnemonics} exemplars being given both \emph{strong optimizability} and \emph{flexible adaptation ability} through the bilevel optimization program (BOP).

\myparagraph{Ablation study.}
Table~\ref{table_ablation} concerns six ablative settings and compares the efficiencies between our \emph{mnemonics} training approach (\emph{w/} and \emph{w/o} adjusting old exemplars) and two baselines: \emph{random} and \emph{herding} exemplars.
Concretely, our approach
achieves the highest average accuracies
and the lowest forgetting rates in all settings.
Dynamically adjusting old exemplars brings consistent improvements, i.e., average $0.34\%$ on both datasets. 
In terms of forgetting rates, our results are the lowest (best). It is interesting that \emph{random} achieves lower (better) performance than \emph{herding}.
\emph{Random} selects exemplars both on the center and boundary of the data space (for each class), but \emph{herding} considers the center data only which strongly relies on the data distribution in the current phase but can not take any risk of distribution change in subsequent phases.
This weakness is further revealed through the visualization of exemplars in the data space, e.g., in Figure~\ref{fig_5phase}. 
Note that the results of ablative study on other components, e.g. distillation loss, are given in the supplementary.

\setlength{\tabcolsep}{1.00mm}{
\begin{table}[htp]
  \small
  \centering
  \begin{tabular}{lcccccccc}
  \toprule
  \multirow{2.5}{*}{Exemplar} & & \multicolumn{3}{c}{\emph{CIFAR-100}} && \multicolumn{3}{c}{\emph{ImagNet-Subset}} \\
   \cmidrule{3-5} \cmidrule{7-9} 
   &&  $N$=5 & 10  & 25 && 5 & 10 & 25  \\
   \midrule
    \emph{random} \emph{w/o} adj. & \multirow{6}{*}{$\uparrow$} & 61.87 & 60.23 & 58.57 && 70.67 & 69.15 & 67.17\\
    \emph{random}  & & 62.64 & 60.61 & 58.82 && 70.69 & 69.67 & 67.46\\
    \emph{herding} \emph{w/o} adj. && 62.98 & 61.23 & 60.36 && 71.66 & 71.02 & 69.40\\
    \emph{herding}  && 62.96 & 61.76 & 60.38 && 71.76 & 71.04 & 69.61\\
    ours \emph{w/o} adj.  && 63.25 & 61.86 & 60.46 && 71.91 & 71.08 & 69.68\\
    ours && {\textbf{63.34}} & {\textbf{62.28}} &  {\textbf{60.96}} && {\textbf{72.58}} &  {\textbf{71.37}} &  {\textbf{69.74}} \\
   \midrule
    \emph{random}  \emph{w/o} adj.  & \multirow{6}{*}{$\downarrow$} & 11.16 & 13.42 & \textbf{12.26} && 18.92 & 19.56 & 23.64\\
    \emph{random}  & & 12.13 & 14.80 & \textbf{12.26} && 17.92 & 17.91 & 23.60\\
    \emph{herding}  \emph{w/o} adj. && 12.69 & 13.63 & 16.36 && 17.16 & 18.00 & \textbf{20.00}\\
    \emph{herding}  && 11.00 & 14.38 & 15.60 && \textbf{15.80} & 17.84 & 20.72\\
    ours \emph{w/o} adj.   && \textbf{9.80} & 13.44 & 16.68 && 18.27 & 18.08 & 20.96\\
    ours && {\redt{10.91}} &  {\textbf{13.38}} &  {{15.22}} && {{17.40}} &  {\textbf{17.08}} &  {\redt{20.83}} \\
  \bottomrule
\end{tabular}
  \cotronlcaptionvsapce
  \caption{Ablation study. The top and the bottom blocks present average accuracies $\bar{\mathcal{A}}$ (\%) and forgetting rates $\mathcal{F}$ (\%), respectively. 
  ``\emph{w/o} adj.'' means without old exemplar adjustment.
  Note that the weight transfer operations are applied in all these experiments.}
  \label{table_ablation}
  \cotronlvsapce
\end{table}
}

\myparagraph{Visualization results.}
Figure~\ref{fig_5phase} demonstrates the t-SNE
results for \emph{herding} (deep-colored) and our \emph{mnemonics} exemplars (deep-colored) in the data space (light-colored).
We have two main observations. (1)~Our \emph{mnemonics} approach results in much clearer separation in the data than \emph{herding}. 
(2)~Our \emph{mnemonics} exemplars are optimized to mostly locate on the boundaries between classes, which is essential to yield high-quality classifiers.
Comparing the Phase-$4$ results of two datasets (i.e., among the sub-figures on the rightmost column), we can see that learning more classes (i.e., on the ImageNet) clearly causes more confusion among classes in the data space, while our approach is able to yield stronger intra-class compactness and inter-class separation. In the supplementary, we present more visualization figures about the changes during the \emph{mnemonics} training from initial examples to learned exemplars.

\section{Conclusions}

In this paper, we develop a novel \emph{mnemonics} training framework for tackling multi-class incremental learning tasks. 
Our main contribution is the \emph{mnemonics} exemplars which are not only efficient data samples but also flexible, optimizable and adaptable parameters contributing a lot to the flexibility of online systems.
Quite intriguingly, our \emph{mnemonics} training approach is \emph{generic} that it can be easily applied to existing methods to achieve large-margin improvements.
Extensive experimental results on four different baseline architectures validate the high efficiency of our approach, and the in-depth visualization reveals the essential reason is that our \emph{mnemonics} exemplars are automatically learned to be the optimal replacement of the original data which can yield high-quality classification models.
\section*{Acknowledgments}
We would like to thank all reviewers for their constructive comments. This research was supported by the Singapore Ministry of Education (MOE) Academic Research Fund (AcRF) Tier 1 grant, Alibaba Innovative Research (AIR) programme, Max Planck Institute for Informatics, the National Natural Science Foundation of China (61772359, 61572356, 61872267), the grant of Tianjin New Generation Artificial Intelligence Major Program (19ZXZNGX00110, 18ZXZNGX00150), the Open Project Program of the State Key Lab of CAD and CG, Zhejiang University (Grant No. A2005), and the grant of Elite Scholar Program of Tianjin University (2019XRX-0035).

{\small
\bibliographystyle{ieee_fullname}
\bibliography{egbib}
}

\clearpage
\beginsupp
\setcounter{section}{0}
\renewcommand\thesection{\Alph{section}}
\noindent
{\Large {\textbf{Supplementary materials}}}
\\

These supplementary materials include the experiments in uniform memory budget setting (\S\ref{suppsec_fixed_budget}), the ablation study for distillation loss (\S\ref{suppsec_distill}), and more visualization results (\S\ref{suppsec_visual}). 

\section{Uniform memory budget experiments}
\label{suppsec_fixed_budget}
\myparagraphsupp{This is supplementary to Section~\textcolor{red}{5.1} ``\textbf{the architecture of $\mathcal{E}$}''.} As mentioned in the main paper,
we have two methods of setting memory budgets~\cite{hou2019learning}. (1) \textbf{Uniform exemplar number setting (used in the main paper)}: every class has $20$ exemplars. 
(2) \textbf{Uniform memory budget setting}: the system works on a uniform memory budget, e.g. $2,000$ ($20,000$) total exemplars can be stored in each phase. In this setting, the system stores more exemplars per class in earlier phases and discards some of the old exemplars afterwards.

\begin{figure*}
\newcommand{\newincludegraphics}[1]{\includegraphics[height=1.26in]{#1}}
\centering
\includegraphics[height=0.16in]{files/legend.pdf}
\vspace{3mm}
\subfigure[CIFAR-100 ($100$ classes). In the $0$-th phase, $\Theta_0$ is trained on $50$ classes, the remaining  classes are given evenly in the subsequent phases.]{
\newincludegraphics{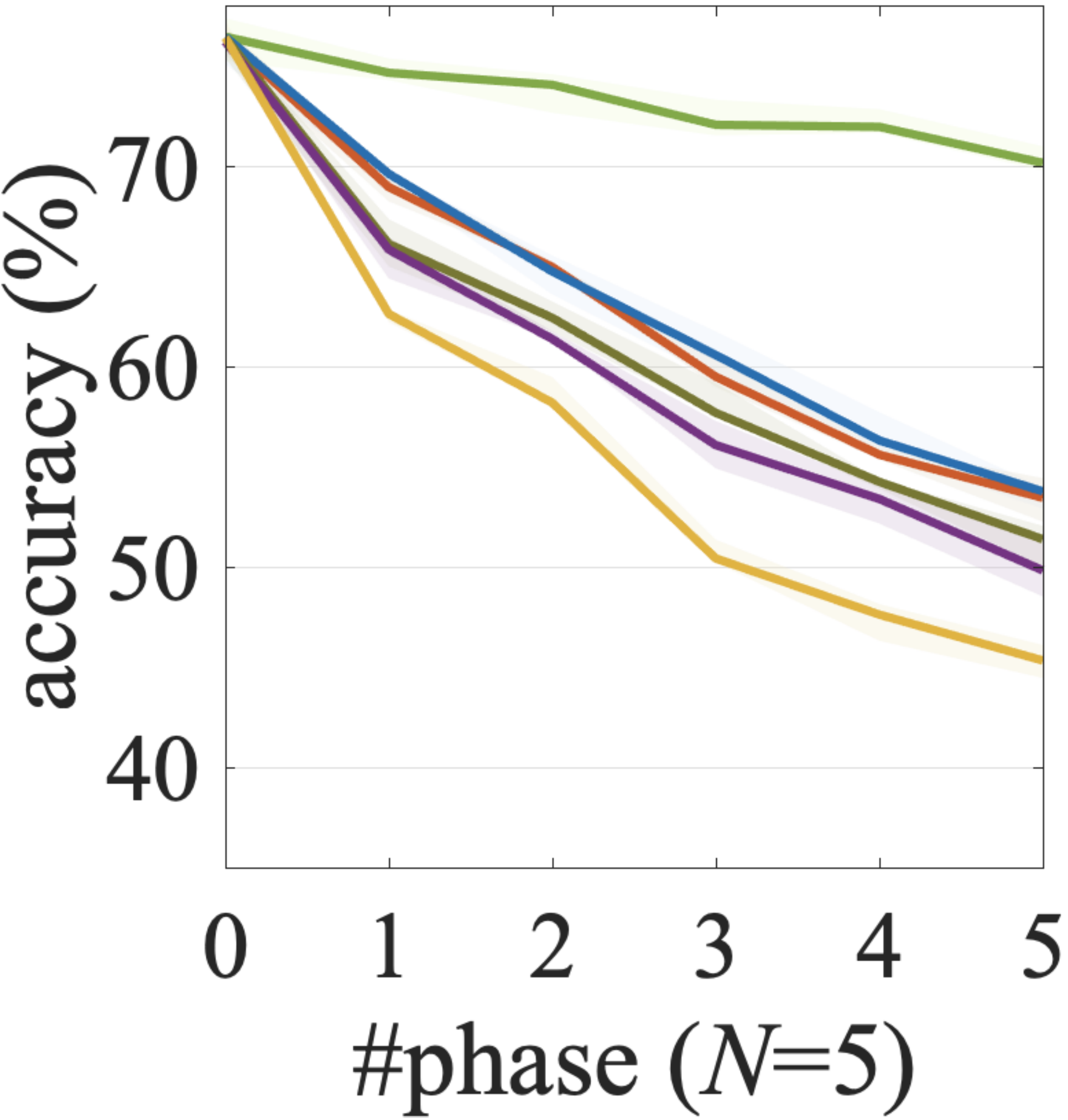}

\hspace{1mm}
\newincludegraphics{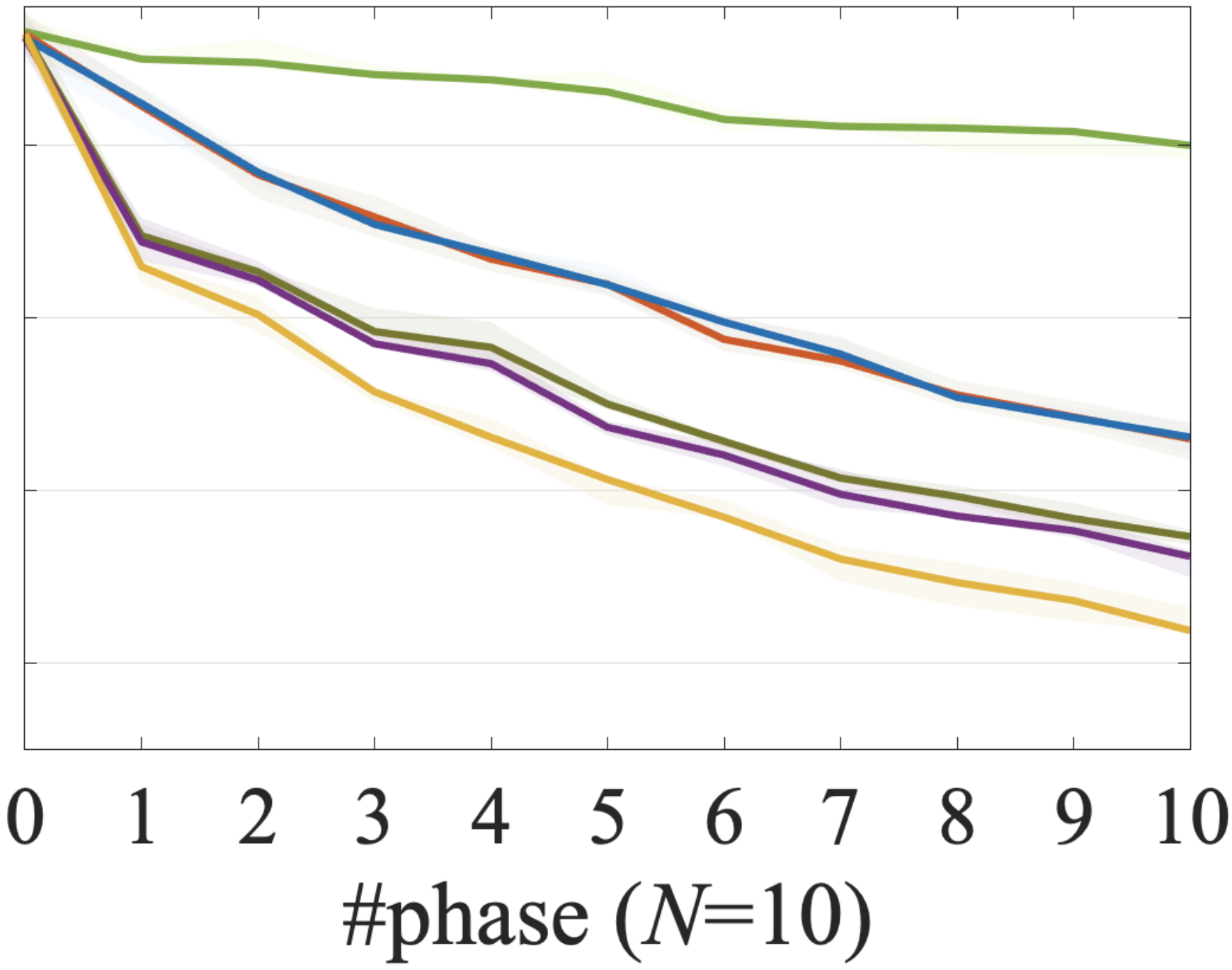}

\hspace{1mm}
\newincludegraphics{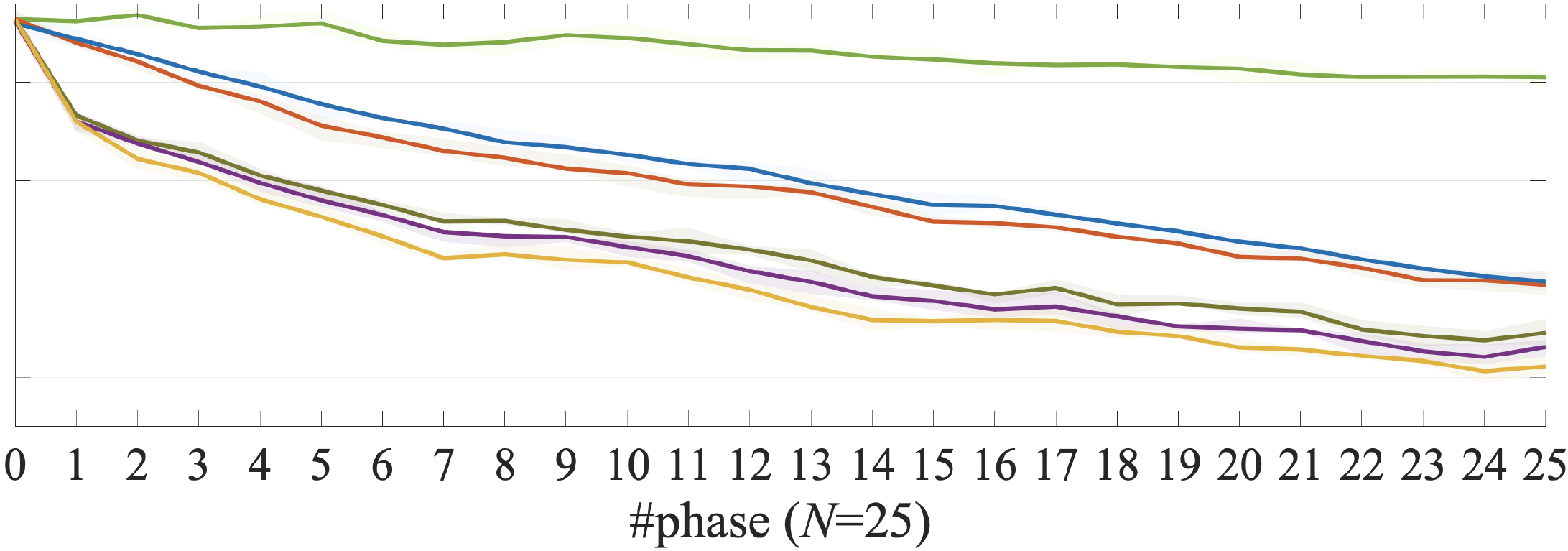}
}
\\
\subfigure[ImageNet-Subset ($100$ classes). In the $0$-th phase, $\Theta_0$ is trained on $50$ classes, the remaining classes are given evenly in the subsequent phases.]{
\newincludegraphics{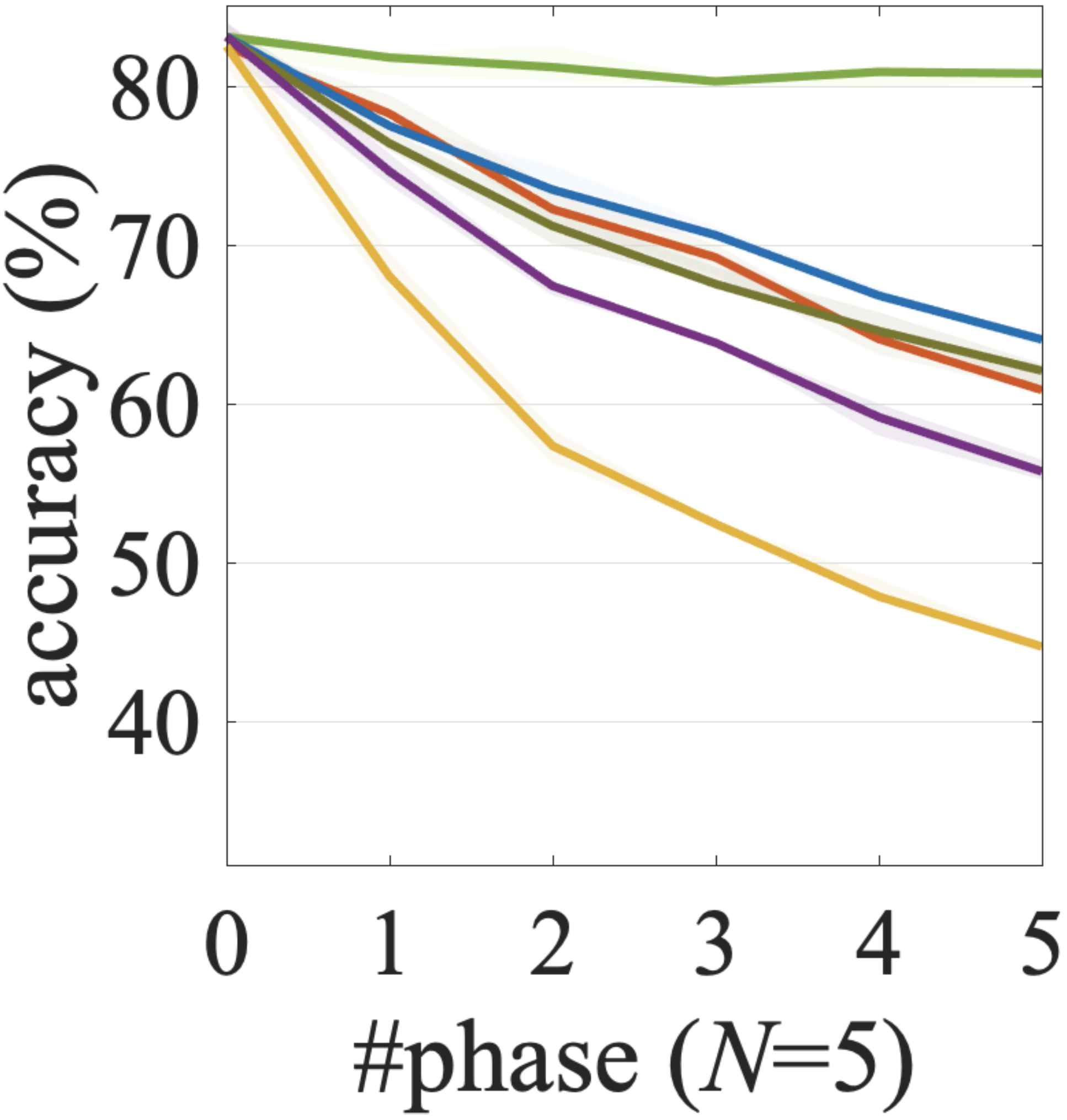}

\hspace{1mm}
\newincludegraphics{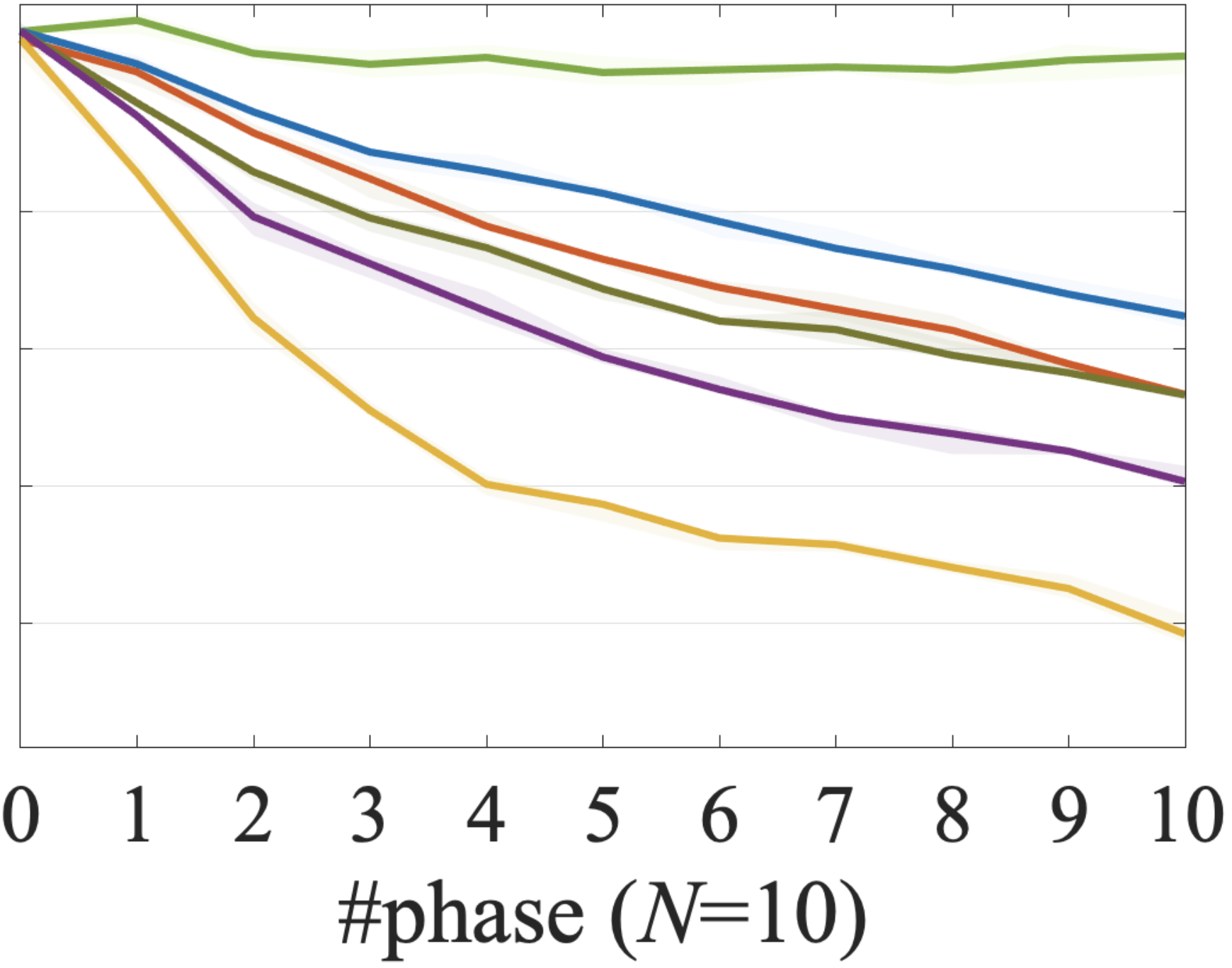}

\hspace{1mm}
\newincludegraphics{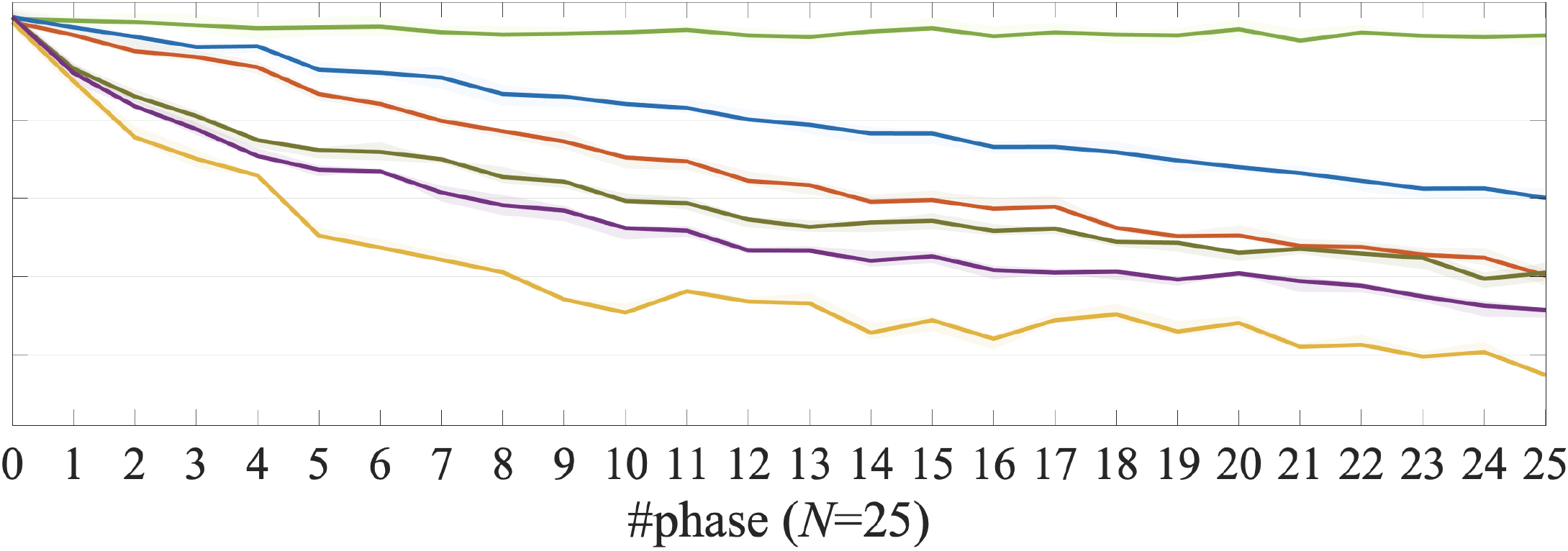}
}
\\
\cotronlcaptionvsapce
\caption{\mycaptionsupp{Supplementary to Figure~\textcolor{red}{4}.} \textbf{Uniform memory budget setting.} Phase-wise accuracies ($\%$).
Light-color ribbons are visualized to show the $95\%$ confidence intervals. 
Comparing methods: Upper Bound (the results of joint training with all previous data accessible in each phase); LUCIR (2019)~\cite{hou2019learning}; BiC (2019)~\cite{Wu2019LargeScale}; iCaRL (2017)~\cite{rebuffi2017icarl}; and LwF (2016)~\cite{Li18LWF}. 
For ours, we show our results using ``LUCIR \emph{w/} ours''. The average accuracy of each curve is given in Table~\ref{table_sota_supp}. 
Note that we apply \emph{\textbf{random}} strategy to discard old exemplars.
}
\label{figure_acc_plots_supp}
\cotronlvsapce
\end{figure*}

In this section, we show the results of \textbf{Uniform memory budget setting}.
Table~\ref{table_sota_supp} presents the 
comparisons to related works~\cite{hou2019learning, Li18LWF, rebuffi2017icarl, Wu2019LargeScale}, with and without our \emph{mnemonics} training as plug-in module. Figure~\ref{figure_acc_plots_supp} demonstrates the phase-wise results of our best model.

\section{Ablation study for distillation loss}
\label{suppsec_distill}

\myparagraphsupp{This is supplementary to Section~\textcolor{red}{5.2} ``\textbf{ablation study}''.}
In Table~\ref{table_ablation_distill}, we supplement the ablation study of distillation loss~\cite{Li18LWF,rebuffi2017icarl}.
\setlength{\tabcolsep}{1.25mm}{
\begin{table}[htp]

  \small
  \centering
  \begin{tabular}{lcccccccc}
  \toprule
  \multirow{2.5}{*}{Exemplar} & & \multicolumn{3}{c}{\emph{CIFAR-100}} && \multicolumn{3}{c}{\emph{ImagNet-Subset}} \\
   \cmidrule{3-5} \cmidrule{7-9} 
   &&  $N$=5 & 10  & 25 && 5 & 10 & 25  \\
   \midrule
    \emph{w/o} distill.  & \multirow{2}{*}{$\uparrow$} & 50.86 & 50.40 & 49.69 && 65.74 & 64.63 & 65.21\\
    \emph{w/} distill. && \textbf{63.34} & \textbf{62.28} & \textbf{60.96} && \textbf{72.58} & \textbf{71.37} & \textbf{69.74} \\
   \midrule
    \emph{w/o} distill.  & \multirow{2}{*}{$\downarrow$}  & 38.70 & 36.94 & 34.38 && 29.32 & 27.44 & 29.64\\
    \emph{w/} distill.  && \textbf{10.91} & \textbf{13.38} & \textbf{19.80} && \textbf{17.40} & \textbf{17.08} & \textbf{20.83} \\
  \bottomrule
\end{tabular}
  \vspace{0.4cm}
  \caption{\mycaptionsupp{Supplementary to Table~\textcolor{red}{2}.} Ablation study for distillation loss. The top and the bottom blocks present average accuracies $\bar{\mathcal{A}}$ (\%) and forgetting rates $\mathcal{F}$ (\%) of using LUCIR~\cite{hou2019learning} with our \emph{mnemonics} training approach or \cite{shin2017continual} as a plug-in module. 
  ``\emph{w/}'' and ``\emph{w/o} distill.'' mean with and without distillation loss, respectively. Note that the weight transfer operations are applied in all these experiments.}
  \label{table_ablation_distill}
  \vspace{-0.3cm}

\end{table}
}

\section{More visualization results}
\label{suppsec_visual}

\begin{figure*}
\newcommand{\newincludegraphics}[1]{\includegraphics[width=2.26in]{#1}}
\newcommand{\newincludegraphicseuc}[1]{\includegraphics[width=2.26in]{#1}}
\centering
\subfigure[CIFAR-100 ($100$ classes). Cosine distance.]{
\newincludegraphics{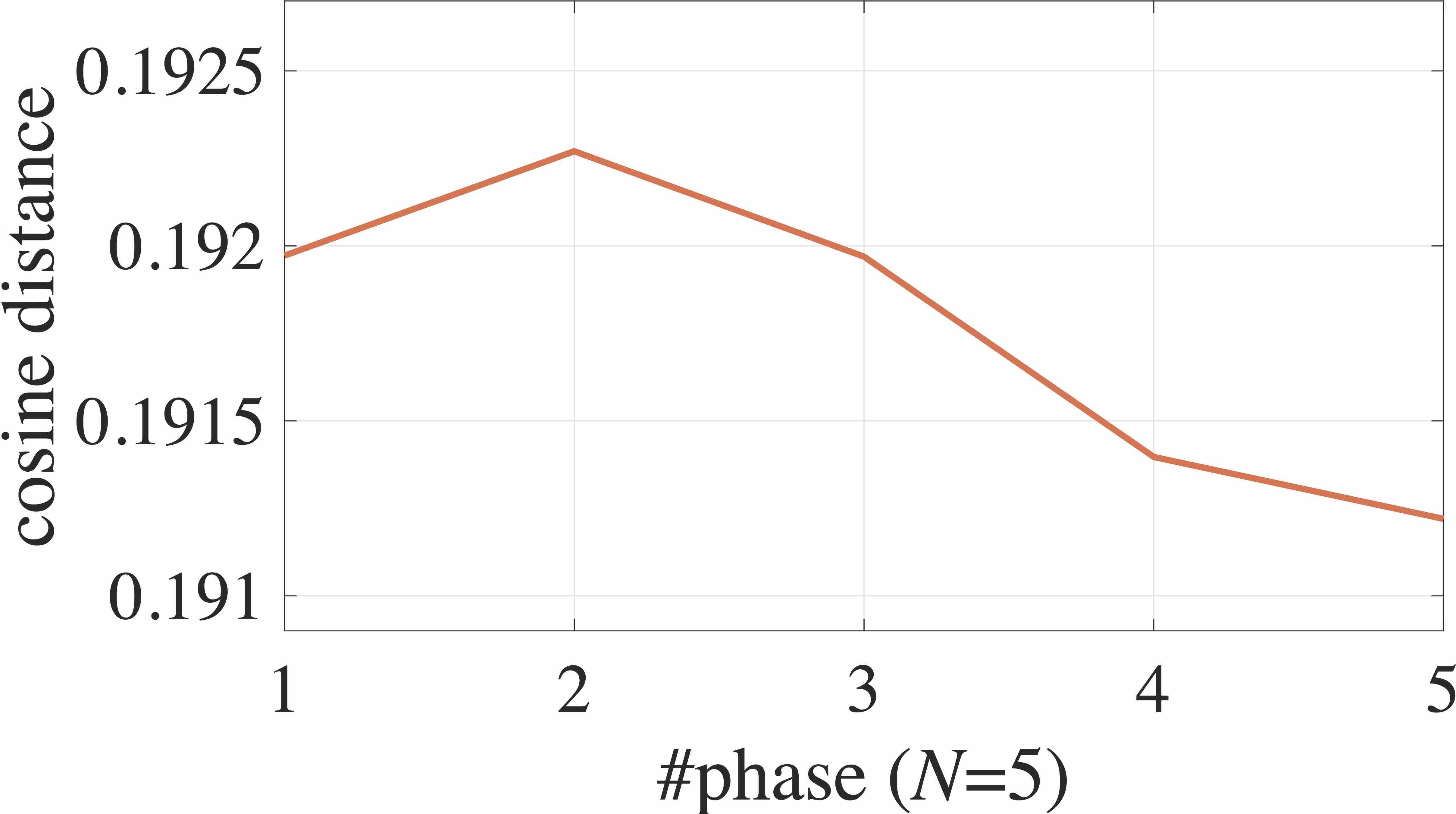}

\hspace{1mm}
\newincludegraphics{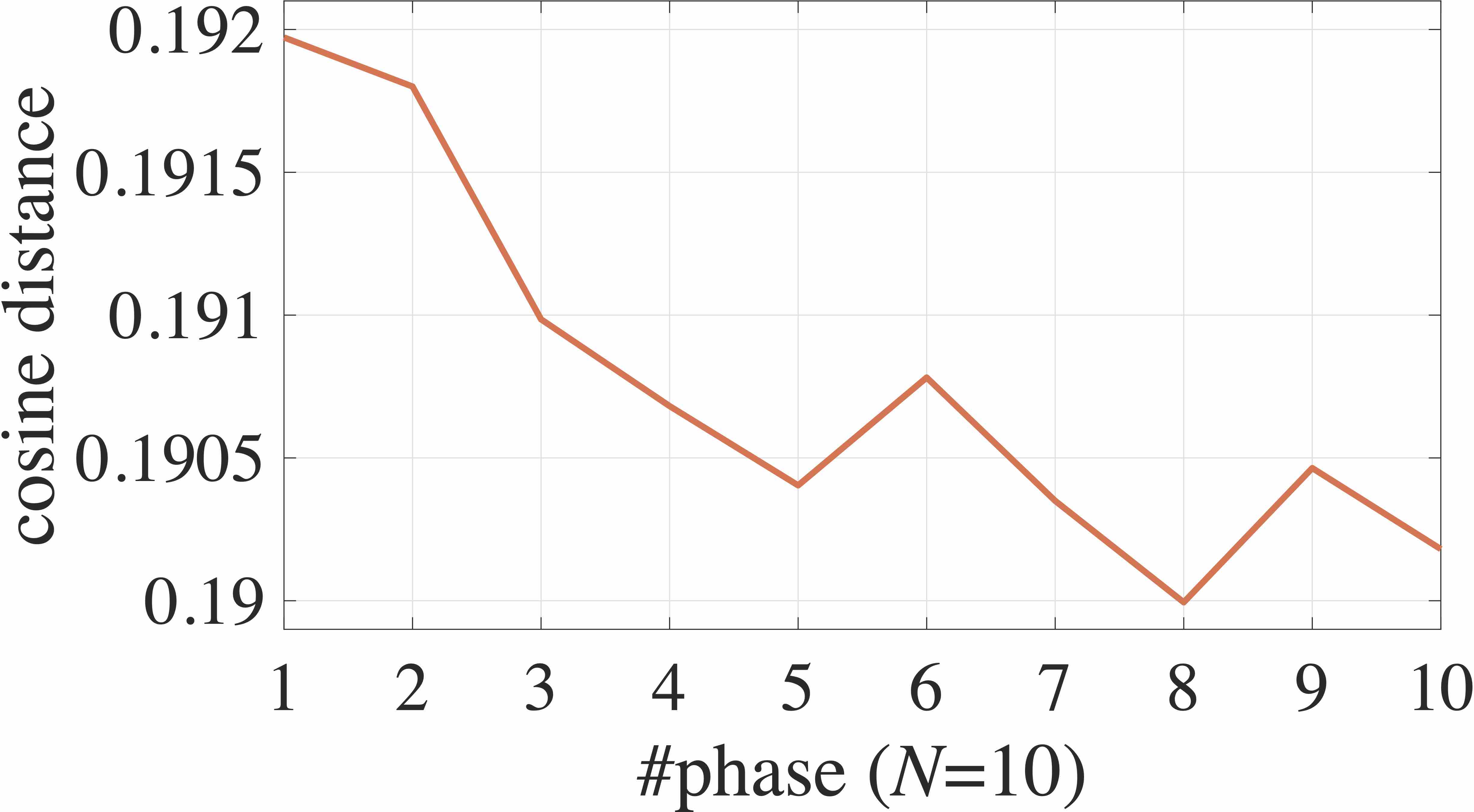}

\hspace{1mm}
\newincludegraphics{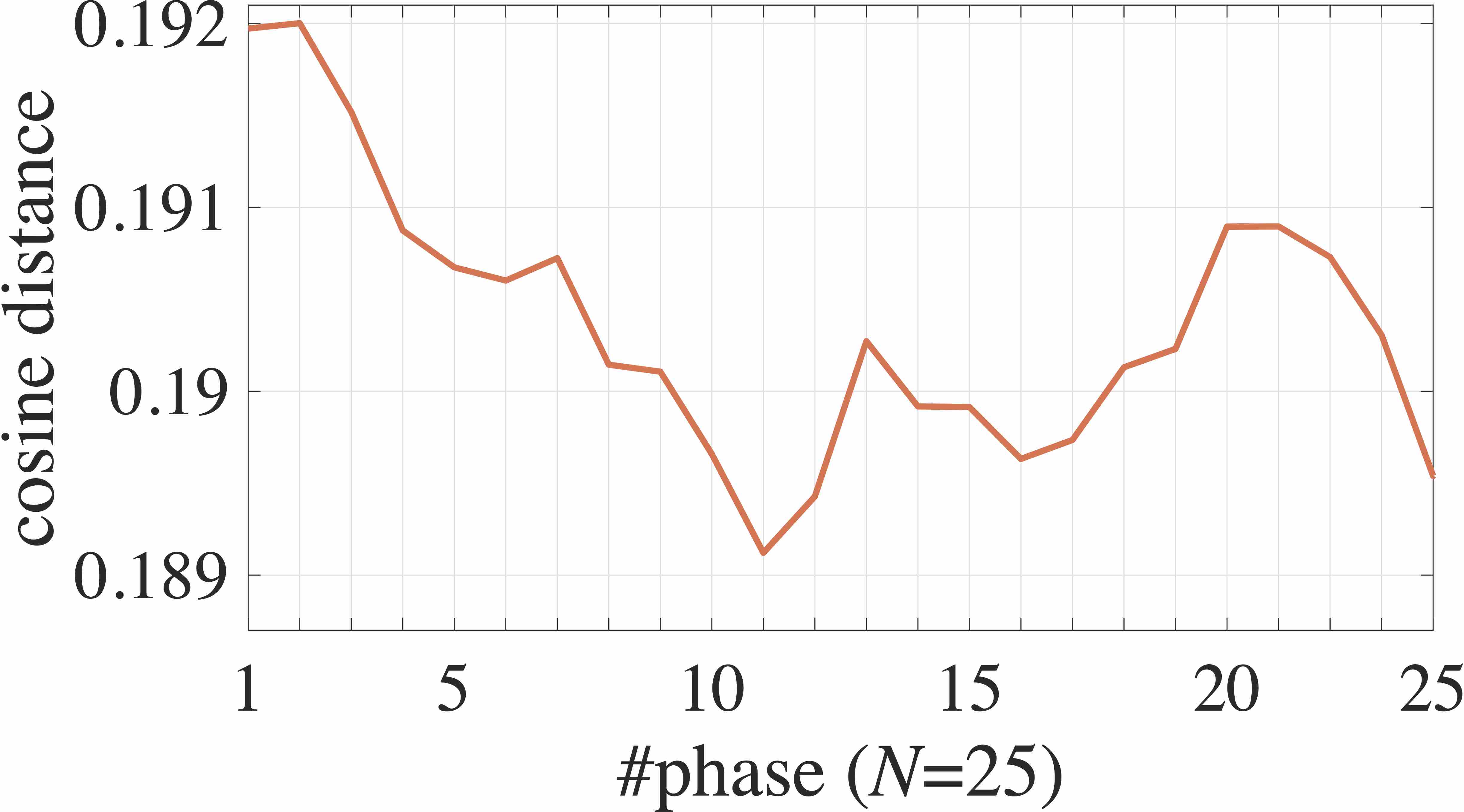}

}
\\
\subfigure[CIFAR-100 ($100$ classes). Euclidean distance.]{
\newincludegraphicseuc{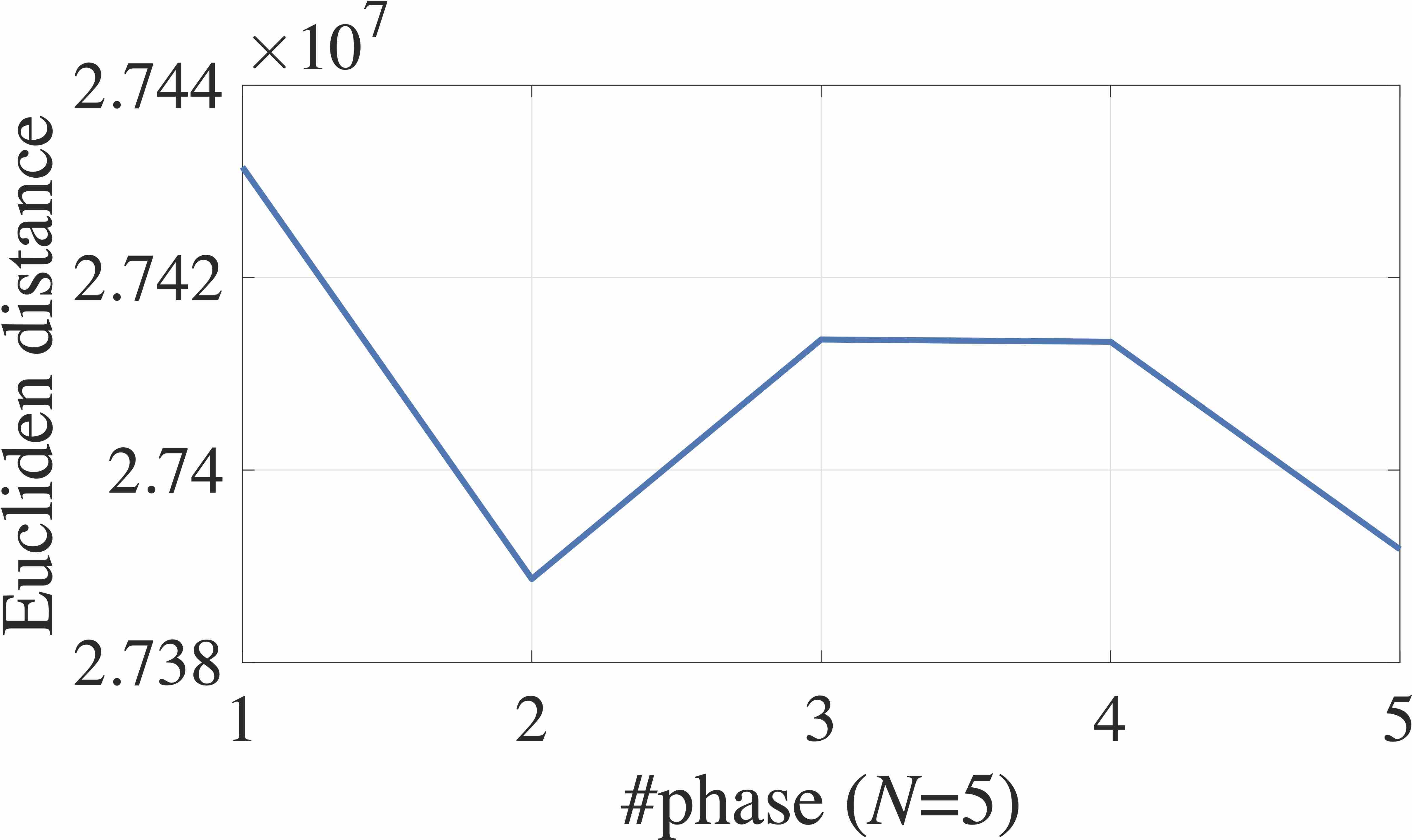}

\hspace{1mm}
\newincludegraphicseuc{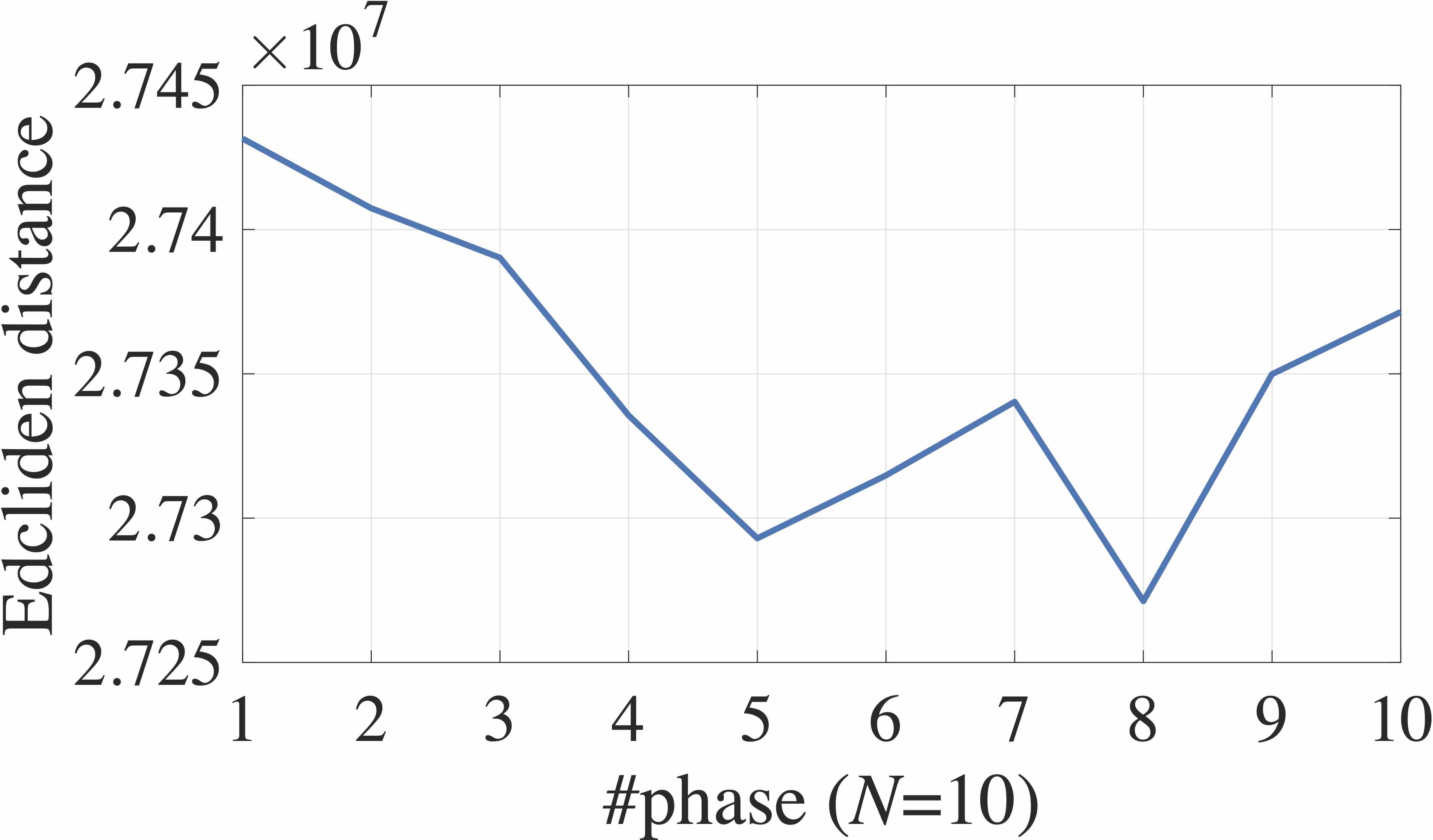}

\hspace{1mm}
\newincludegraphicseuc{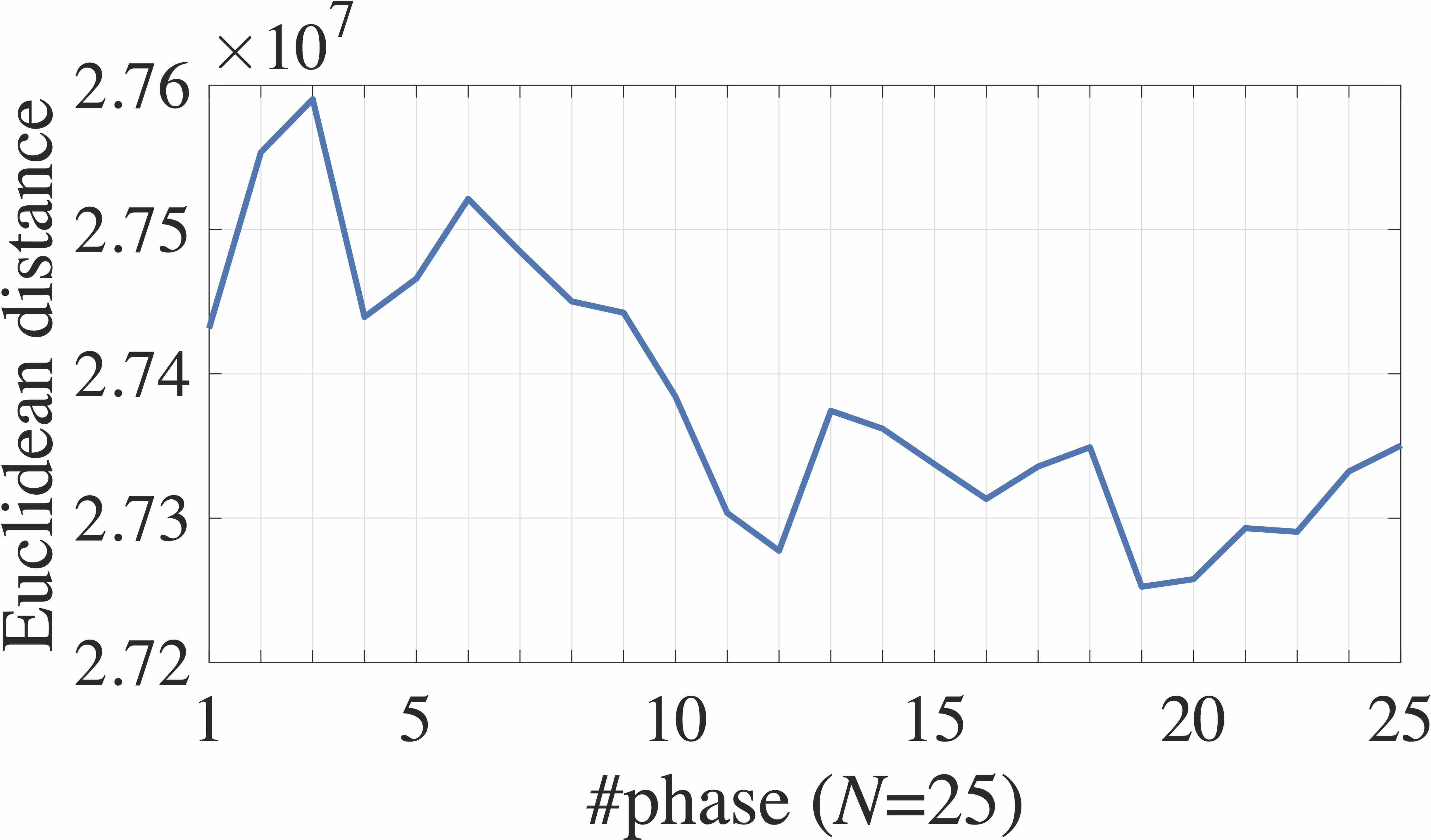}

}
\\
\cotronlcaptionvsapce
\caption{The changes of average distances between exemplars and initial samples. We show the results using ``LUCIR \emph{w/} ours''. The average accuracy of each curve is given in Table~\redt{1}. All curves are smoothed with a rate of $0.8$ for a better visualization. 
}
\label{figure_distance}
\cotronlvsapce
\end{figure*}

\myparagraphsupp{This is supplementary to Section~\textcolor{red}{5.2} ``\textbf{visualization results.}''.}
In Figure~\ref{figure_distance}, we supplement the changes of average distances between exemplars and initial samples..

\setlength{\tabcolsep}{1.10mm}{\begin{table}[htb]
  \small
  \centering
  \begin{tabular}{llccccccc}
  \toprule
  \multirow{2.5}{*}{Method} && \multicolumn{3}{c}{\emph{CIFAR-100}} && \multicolumn{3}{c}{\emph{ImageNet-Subset}} \\
  \cmidrule{3-5} \cmidrule{7-9}
   && $N$=5 & 10  & 25 && 5 & 10 & 25  \\
   \midrule
   LwF${}^{\diamond}$~\cite{Li18LWF}& \multirow{9.5}{*}{$\uparrow$} & 56.79 & 53.05 & 50.44 && 58.83 & 53.60 & 50.16 \\
   LwF \emph{w/} ours  && 57.64 & 56.43 & 55.42 && 64.14 & 62.41 & 61.51  \\
   \cmidrule{3-9}
   iCaRL~\cite{rebuffi2017icarl}  && 60.48 & 56.04 & 52.07 && 67.32 & 62.42 & 57.04  \\
   iCaRL \emph{w/} ours & & 61.78 & 59.56 & 56.80 && \textbf{73.17} & 71.33 & 67.80  \\
   \cmidrule{3-9}
   BiC~\cite{Wu2019LargeScale} & & 61.35 & 56.81 & 53.42 && 70.82 & 66.62 & 60.44  \\
   BiC \emph{w/} ours  && 62.28 & 60.31 & 57.85 && 73.05 & 71.60 & 68.70  \\
   \cmidrule{3-9}
   LUCIR~\cite{hou2019learning}   && 63.34 & 62.47 & 59.69 && 71.20 & 68.21 & 64.15  \\
   LUCIR \emph{w/} ours  && \textbf{64.58} & \textbf{62.58} & \textbf{61.02} && {72.60} & \textbf{71.66} & \textbf{70.51}  \\
  \midrule
   \midrule
   LwF${}^{\diamond}$~\cite{Li18LWF}  & \multirow{9.5}{*}{$\downarrow$} & 41.07 & 40.20 & 38.64 && 52.48 & 53.20 & 51.16  \\
   LwF \emph{w/} ours  && 36.14 & 34.28 & 32.92 && 38.00 & 37.64 & 39.63  \\
   \cmidrule{3-9}
   iCaRL~\cite{rebuffi2017icarl}  && 29.42 & 30.80 & 33.32 && 40.40 & 41.76 & 44.04  \\
   iCaRL \emph{w/} ours & & 23.22 & 24.72 & 27.74 && 20.40 & 22.48 & 31.04  \\
   \cmidrule{3-9} 
   BiC~\cite{Wu2019LargeScale}  && 24.96 & 27.14 & 30.66 && 25.56 & 29.40 & 35.84  \\
   BiC \emph{w/} ours && 21.80 & 21.66 & 25.54 && \textbf{14.32} & 17.72 & 25.64  \\
   \cmidrule{3-9}   
   LUCIR~\cite{hou2019learning}  && 20.46 & 23.16 & 25.76 && 27.96 & 31.04 & 35.36  \\
   LUCIR \emph{w/} ours & & \textbf{10.18} & \textbf{11.06} & \textbf{16.04} && {16.56} & \textbf{13.80} & \textbf{21.96}  \\
  \bottomrule

    \multicolumn{8}{l}{${}^{\diamond}$ Using \emph{herding} exemplars as~\cite{hou2019learning,rebuffi2017icarl,Wu2019LargeScale} for fair comparison.}\\
\end{tabular}
  \vspace{0.2cm}
  \cotronlcaptionvsapce
  \caption{ \mycaptionsupp{Supplementary to Table~\textcolor{red}{1}.} 
  \textbf{Uniform memory budget setting.}
  Average accuracies $\bar{\mathcal{A}}$ (\%) (top block) and forgetting rates $\mathcal{F}$ (\%) (bottom block) of using related methods ~\cite{hou2019learning, Li18LWF,rebuffi2017icarl,Wu2019LargeScale} with and without our \emph{mnemonics} training approach as a plug-in module. 
  We use \emph{\textbf{random}} strategy to discard old exemplars, and use weight transfer operations to train MCIL models.
  }
  \label{table_sota_supp}
  \cotronlvsapce
\end{table}}

\end{document}